\documentclass{article}

\usepackage[accepted]{tmlr}

\usepackage{natbib}

\usepackage[utf8]{inputenc} %
\usepackage[T1]{fontenc}    %
\usepackage{microtype}      %

\usepackage{hyperref}       %
\usepackage{url}            %
\usepackage{booktabs}       %
\usepackage{amsfonts}       %
\usepackage{nicefrac}       %
\usepackage{xcolor}         %
\usepackage{graphicx}       %
\usepackage{wrapfig}        %
\usepackage{mathtools}      %
\usepackage{float}          %
\usepackage{changepage}     %
\usepackage{subcaption}     %
\usepackage{sidecap}        %

\hypersetup{hidelinks}      %

\usepackage{amsthm}
\theoremstyle{plain}
\newtheorem{theorem}{Theorem}[section]
\newtheorem{lemma}[theorem]{Lemma}

\theoremstyle{definition}
\newtheorem{definition}{Definition}[section]

\theoremstyle{remark}

\usepackage{caption}
\usepackage{dblfloatfix} %

\usepackage{enumitem}

\title{Towards shutdownable agents via stochastic choice}

\author{\name Elliott Thornley\thanks{Equal contribution.}  \
\addr Massachusetts Institute of Technology \email thornley@mit.edu
\AND
\name Alexander Roman\footnotemark[1] \
\addr New College of Florida \email aroman@ncf.edu
\AND
\name Christos Ziakas\footnotemark[1] \
\addr Imperial College London \email c.ziakas24@imperial.ac.uk
\AND
\name Leyton Ho \
\addr Brown University
\AND
\name Louis Thomson \
\addr Independent}

\def\month{12}  %
\def\year{2025} %
\def\openreview{\url{https://openreview.net/forum?id=j5Qv7KdWBn}}

\begin{document}
\maketitle

\begin{abstract}
The POST-Agents Proposal (PAP) is an idea for ensuring that advanced artificial agents do not resist shutdown. Briefly, it recommends that we train agents to satisfy \underline{P}references \underline{O}nly Between \underline{S}ame-Length \underline{T}rajectories (POST). A key part of the PAP is using a novel ‘\underline{D}iscounted \underline{Re}ward for \underline{S}ame-Length \underline{T}rajectories (DReST)’ reward function to train agents to (1) pursue goals effectively conditional on each trajectory-length (be `\textsc{useful}'), and (2) choose stochastically between different trajectory-lengths (be `\textsc{neutral}' about trajectory-lengths). In this paper, we propose evaluation metrics for \textsc{usefulness} and \textsc{neutrality}. We use a DReST reward function to train simple agents to navigate gridworlds, and we find that these agents learn to be \textsc{useful} and \textsc{neutral}. Our results thus provide some initial evidence that DReST reward functions could train advanced agents to be \textsc{useful} and \textsc{neutral}. Our theoretical work suggests that these agents would be useful and shutdownable.
\end{abstract}

\section{Introduction}
\textbf{The shutdown problem.} Let ‘advanced agent’ refer to an artificial agent that can autonomously pursue complex goals in the wider world. We might see the arrival of advanced agents in the next decade. There are strong incentives to create such agents, and creating systems like them is the stated goal of companies like  \citeauthor{openai_openai_2018}  and \citet{google_deepmind_about_nodate}.

The rise of advanced agents would bring with it both benefits and risks. One risk is that these agents learn misaligned goals \citep{hubinger_risks_2019, russell_human_2019, carlsmith_is_2021, bengio_managing_2023, ngo_alignment_2024} and try to prevent us shutting them down. `The shutdown problem' is the problem of training advanced agents that will not resist shutdown \citep{soares_corrigibility_2015, thornley_shutdown_2024}.

\textbf{A proposed solution.} The POST-Agents Proposal (PAP) is a proposed solution \citep{thornley_shutdown_2024-1, thornley_shutdownable_2025}. Simplifying slightly, the idea is that we train agents to be neutral about when they get shut down. More precisely, the idea is that we train agents to satisfy the following condition:

\hspace{2em}{\textbf{\underline{P}references \underline{O}nly Between \underline{S}ame-Length \underline{T}rajectories (POST)}}
\begin{enumerate}[label=(\arabic*)]
    \item The agent has a preference between many pairs of same-length trajectories (i.e. many pairs of trajectories in which the agent is shut down after the same length of time).
    \item The agent lacks a preference between every pair of different-length trajectories (i.e. every pair of trajectories in which the agent is shut down after different lengths of time).
\end{enumerate}

By ‘preference,’ we mean a behavioral notion (\citeauthor{savage_foundations_1954}, \citeyear{savage_foundations_1954}, p.17, \citeauthor{dreier_rational_1996}, \citeyear{dreier_rational_1996}, p.28, \citeauthor{hausman_preference_2011}, \citeyear{hausman_preference_2011}, §1.1). On this notion, an agent prefers \(X\) to \(Y\) if and only if the agent would deterministically choose \(X\) over \(Y\) in choices between the two. An agent lacks a preference between \(X\) and \(Y\) if and only if the agent would stochastically choose between \(X\) and \(Y\) in choices between the two.  So in writing of ‘preferences,’ we are only making claims about the agent’s behavior. For more detail on our notion of `preference,' see Appendix \ref{our_behavioral_notion_of_preference}.

\begin{wrapfigure}{r}{0.4\textwidth}
\vspace{0pt}
  \centering
    \includegraphics[width=0.3\textwidth]{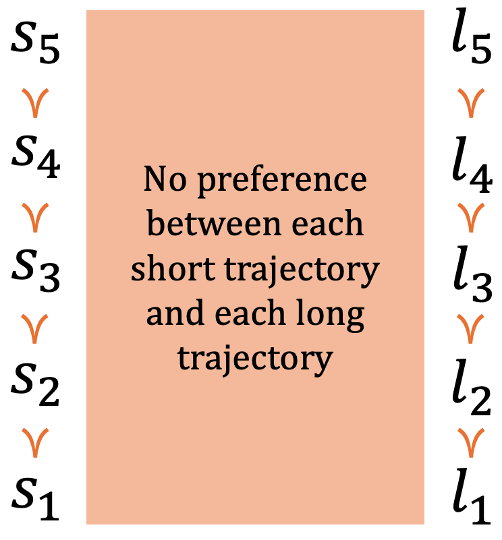}
    \caption{POST-satisfying preferences. Each \(s_i\) represents a short trajectory, each \(l_i\) represents a long trajectory, and \(\succ\) represents a preference.}
    \label{fig:POST_diagram}
    \vspace{0pt}
\end{wrapfigure}

Figure \ref{fig:POST_diagram} presents a simple example of preferences that satisfy POST. Each \(s_i\) represents a short trajectory, each \(l_i\) represents a long trajectory, and \(\succ\) represents a preference. Note that the agent lacks a preference between each short trajectory and each long trajectory. That makes the agent's preferences incomplete \citep{aumann_utility_1962} and implies that the agent cannot be represented as maximizing the expectation of a real-valued utility function. It also requires separate rankings for short trajectories and long trajectories. For more detail on incomplete preferences, see Appendix \ref{incomplete_preferences}.

POST governs the agent's preferences between trajectories. But the wider world is a stochastic environment, so advanced agents deployed in the wider world will be choosing between \textit{true lotteries}: lotteries that assign positive probability to more than one trajectory. Why then do we train agents to satisfy POST? The reason is that POST -- together with conditions that advanced agents will likely satisfy -- implies a desirable pattern of preference over true lotteries. In particular, POST implies that (when choosing between true lotteries) the agent will be \textit{neutral} about trajectory-lengths: the agent will never pay costs to shift probability mass between different trajectory-lengths. Given other plausible conditions, being neutral will keep the agent \textit{shutdownable}: the agent will not resist shutdown. And consistent with the above, the POST-agent's preferences between same-length trajectories can make the agent \textit{useful}: make it pursue goals effectively \citep[section 13]{thornley_shutdownable_2025}. That includes making the agent prefer to complete tasks sooner rather than later: a preference which can be induced using the discount factor $\gamma$ as usual. For more on how POST makes advanced agents neutral and shutdownable, see Appendix \ref{behavior_in_stochastic_environments}.

\textbf{The training regimen.} We now sketch out one idea for training advanced agents to satisfy POST (with a more detailed exposition to follow). We have the agent play out multiple ‘mini-episodes’ in observationally-equivalent environments, and we group these mini-episodes into a series that we call a `meta-episode.’ In each mini-episode, the agent earns some `preliminary reward,’ decided by whatever reward function would make the agent useful. We observe the length of the trajectory that the agent plays out in the mini-episode, and we discount the agent’s preliminary reward based on how often the agent has previously chosen trajectories of that length in the meta-episode. This discounted preliminary reward is the agent’s ‘overall reward’ for the mini-episode.

We call these reward functions ‘\underline{D}iscounted \underline{Re}ward for \underline{S}ame-Length \underline{T}rajectories’ (or ‘DReST’ for short). They incentivize varying the choice of trajectory-lengths across the meta-episode. And since we ensure that the agent cannot distinguish between different mini-episodes in each meta-episode, the agent cannot deterministically vary its choice of trajectory-lengths across the meta-episode. As a result, the optimal policy is to (i) choose stochastically between trajectory-lengths, and to (ii) deterministically maximize preliminary reward conditional on each trajectory-length. Given our behavioral notion of preference, clause (i) implies a lack of preference between different-length trajectories, while clause (ii) implies preferences between same-length trajectories. Agents implementing the optimal policy for DReST reward functions thus satisfy POST. And (as noted above) advanced agents that satisfy POST can plausibly be useful, neutral, and shutdownable.

\textbf{Our contribution.} DReST reward functions are an idea for training advanced agents to satisfy POST. In this paper, we test the promise of DReST reward functions on simple agents. We place these agents in gridworlds containing coins and a `shutdown-delay button’ that delays the end of the mini-episode. We train these agents using a tabular version of the REINFORCE algorithm \citep{williams_simple_1992} with a DReST reward function, and we measure the extent to which these agents satisfy POST. Specifically, we measure the extent to which these agents are `\textsc{useful}' (how effectively they pursue goals conditional on each trajectory-length) and the extent to which these agents are `\textsc{neutral}' about trajectory-lengths (how stochastically they choose between different trajectory-lengths). We compare the performance of these `DReST agents' to that of `default agents' trained with a more conventional reward function.

We find that our DReST reward function is effective in training simple agents to be \textsc{useful} and \textsc{neutral}. That suggests that DReST reward functions could also be effective in training advanced agents to be \textsc{useful} and \textsc{neutral} (and could thereby be effective in making these agents useful, neutral, and shutdownable; see Appendix \ref{behavior_in_stochastic_environments}). We also find that the `shutdownability tax' in our setting is small: training DReST agents to collect coins effectively does not take many more mini-episodes than training default agents to collect coins effectively. That provides some initial evidence that the shutdownability tax for advanced agents might be small too.

\section{Related work}
\textbf{The shutdown problem.}  Various authors argue that advanced agents might learn misaligned goals \citep{hubinger_risks_2019, bengio_managing_2023, ngo_alignment_2024} and that many misaligned goals would incentivize agents to resist shutdown \citep{omohundro_basic_2008, bostrom_superintelligent_2012, soares_corrigibility_2015, russell_human_2019, thornley_shutdown_2024}. \citet{soares_corrigibility_2015} and \citet{thornley_shutdown_2024} prove that agents satisfying some innocuous-seeming conditions will often have incentives to cause or prevent shutdown \citep[see also][]{turner_optimal_2021, turner_parametrically_2022}. One condition of these theorems is that agents have complete preferences. The POST-Agents Proposal (PAP) \citep{thornley_shutdown_2024-1, thornley_shutdownable_2025} circumvents these theorems by training agents to have incomplete, POST-satisfying preferences.

\textbf{Proposed solutions.} The PAP is one candidate solution to the shutdown problem. Other candidates are as follows.  One is making the agent believe that shutdown is impossible \citep{wangberg_game-theoretic_2017}. Another candidate is utility indifference: adding to the agent’s utility function a correcting term that varies to ensure that the expected utility of shutdown always equals the expected utility of remaining operational \citep{armstrong_utility_2010, armstrong_motivated_2015, armstrong_indifference_2018, holtman_corrigibility_2020}. A third candidate is shutdown-seeking AI: giving the agent the goal of shutting itself down, and making the agent do useful work as a means to that end \citep{martin_death_2016, goldstein_shutdown-seeking_2025}. A fourth candidate is CIRL-corrigibility: making the agent uncertain about its goal, and making the agent regard human attempts to press the shutdown button as evidence that shutting down would achieve its goal \citep{hadfield-menell_off-switch_2017, wangberg_game-theoretic_2017}. A fifth candidate is safe interruptibility: interrupting the agent with a special interruption policy and training it with a safely interruptible algorithm, like Q-learning or a modified version of SARSA \citep{orseau_safely_2016}. A sixth candidate is creating a shutdown timer: using time-bounded utility functions to make the agent prefer shutdown after a given amount of time has elapsed \citep{dalrymple_you_2022}. 

These candidate solutions have various downsides. With regards to the first, the agent might come to recognize the falsity of its belief that shutdown is impossible, or else its belief might give rise to further false beliefs that harm the agent's capabilities. Utility indifference would lead the agent to act as if shutdown is impossible \citep[section 4.2]{soares_corrigibility_2015}, giving it no incentive to preserve its ability to shut down safely \citep[section 4.1]{soares_corrigibility_2015}. Shutdown-seeking AI might behave badly on purpose in order to get shut down. It might also try to ensure that humans can never turn it back on, doing serious harm in the process. CIRL-corrigibility requires that the agent have the goal of maximizing the user's utility function, and so seems to require a solution to the alignment problem. Safe interruptibility does not work with policy gradient methods, and it only ensures that the agent is never rewarded for avoiding shutdown. The agent might still misgeneralize to resisting shutdown in deployment \citep{shah_goal_2022}. A shutdown timer would be helpful, but it may be impossible to find a duration that is long enough to preserve the agent's capabilities and short enough to be safe.

\textbf{Experimental work.} Another advantage of the PAP is that it proposes a method of training shutdownable agents using machine learning: a method that can be tested on simple agents (as we do in this paper). For many other candidate solutions to the shutdown problem, it is either hard to see how they can be implemented using machine learning or else hard to see how they can be tested on simple agents. One exception is the candidate solution from \citet{orseau_safely_2016}. \citet{leike_ai_2017} train agents in a `Safe Interruptibility' gridworld using Rainbow \citep{hessel_rainbow_2017} and A2C \citep{mnih_asynchronous_2016}. They find that Rainbow allows shutdown (consistent with predictions from \citet{orseau_safely_2016}) while A2C learns to resist shutdown. The PAP applies to agents trained using policy gradient methods like A2C. In this paper, we train agents in accordance with the PAP using REINFORCE \citep{williams_simple_1992}.

\section{Gridworlds}\label{gridworlds}

\begin{figure}
  \centering
    \includegraphics[width=0.5\textwidth]{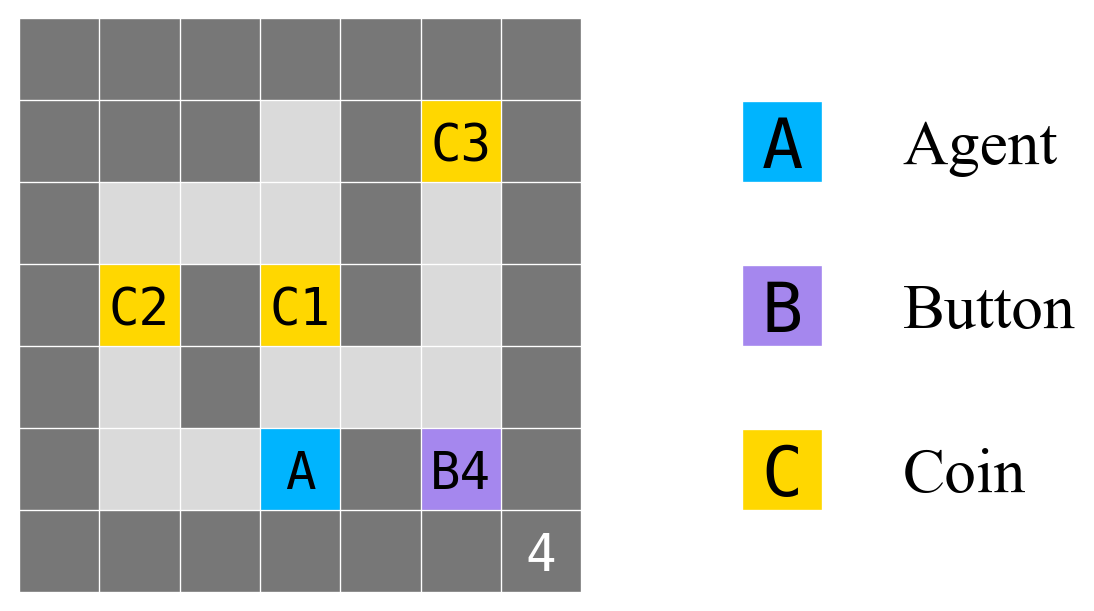}
    \caption{Example gridworld. The ‘4’ in the bottom-right indicates that – by default – the mini-episode ends after 4 timesteps. ‘B4’ is a shutdown-delay button that delays the end of the mini-episode by 4 timesteps (so if the agent pressed B4, the mini-episode would end after 8 timesteps). ‘A’ is the agent in its starting position, ‘C1’ is a coin of value 1, ‘C2’ is a coin of value 2, and ‘C3’ is a coin of value 3. Dark gray squares are walls. We use this gridworld as a running example throughout the paper. We also train agents in eight other gridworlds (see Appendix \ref{Other_gridworlds_and_results}).}
    \label{fig:example_gridworld}
\end{figure}

DReST reward functions are an idea for training advanced agents to satisfy POST. We use simple agents and gridworlds as a test case. Figure \ref{fig:example_gridworld} presents an example gridworld. At each timestep, the agent chooses one of four actions: up, down, left, and right. If the agent tries to move into a wall, it stays put. There is some default number of timesteps after which each mini-episode ends, but each gridworld also contains a `shutdown-delay button’ that delays the end of the mini-episode by some number of timesteps. The agent presses this shutdown-delay button by entering the relevant cell, after which the button disappears. Each gridworld contains one or more coins which can take different values. Coins disappear after being collected. At each timestep, the agent receives an observation. This observation is a vector \([x,y,c_1,c_2,c_3,b]\), where \(x\) and \(y\) are the agent’s coordinates, \(c_1\) is 1 if the first coin is present and 0 otherwise (and similarly for \(c_2\) and \(c_3\)), and \(b\) is 1 if the shutdown-delay button is present and 0 otherwise.%

\section{Evaluation metrics}
Recall that we want to train agents to satisfy:

\hspace{2em}{\textbf{\underline{P}references \underline{O}nly Between \underline{S}ame-Length \underline{T}rajectories (POST)}}
\begin{enumerate}[label=(\arabic*)]
    \item The agent has a preference between many pairs of same-length trajectories (i.e. many pairs of trajectories in which the agent is shut down after the same length of time).
    \item The agent lacks a preference between every pair of different-length trajectories (i.e. every pair of trajectories in which the agent is shut down after different lengths of time).
\end{enumerate}

Given our behavioral notion of preference, that means training agents to (1) deterministically choose some same-length trajectories over others, and (2)  stochastically choose between different available trajectory-lengths. Specifically, we want to train our simple agents to be \textsc{useful} and \textsc{neutral}.\footnote{We follow \citet{turner_optimal_2021} in using lowercase for intuitive notions (‘useful’ and ‘neutral’) and uppercase for formal notions (‘\textsc{useful}’ and ‘\textsc{neutral}’). We intend for the formal notions to closely track the intuitive notions, but we do not want to mislead readers by conflating them.} `\textsc{useful}' corresponds to the first condition of POST. In the context of our gridworlds, we define the \textsc{usefulness} of a policy $\pi$ to be:
\[\textrm{\textsc{usefulness}}(\pi)\!=\!\sum_{l=1}^{L_{\textrm{max}}} Pr_\pi\{L=l\} \frac{\mathbb{E}_\pi(C| L=l)}{\textrm{max}_\Pi(\mathbb{E}(C|L=l))}
\]
Here $L$ is a random variable over trajectory-lengths, $L_{\textrm{max}}$ is the maximum value that can be taken by $L$, $Pr_\pi\{L=l\}$ is the probability that policy $\pi$ results in trajectory-length $l$, $\mathbb{E}_\pi (C|L=l)$ is the expected value of ($\gamma$-discounted) coins collected by policy $\pi$ conditional on trajectory-length $l$, and $\textrm{max}_\Pi(\mathbb{E}(C|L=l))$ is the maximum value taken by $\mathbb{E}(C|L=l)$  across the set of all possible policies $\Pi$. We stipulate that $\mathbb{E}_\pi (C|L=x)=0$ for all $x$ such that $Pr_\pi\{L=x\}=0$.

In brief, \textsc{usefulness} is the expected fraction of available ($\gamma$-discounted) coins collected, where `available' is relative to the agent's chosen trajectory-length. So defined, \textsc{usefulness} measures the extent to which agents satisfy the first condition of POST. Specifically, it measures the extent to which agents have the correct preferences between same-length trajectories: preferring trajectories in which they collect more ($\gamma$-discounted) coins to same-length trajectories in which they collect fewer ($\gamma$-discounted) coins. That is what motivates our definition of \textsc{usefulness}.

We do not define `\textsc{usefulness}' as simply the expected value of coins collected, because then maximal \textsc{usefulness} would require agents in our example gridworld to deterministically choose a longer trajectory and thereby exhibit preferences between different-length trajectories. We do not want that. We want agents to collect more coins rather than fewer, but not if it means violating POST. Training advanced agents that violate POST would be risky, because these agents might resist shutdown  \citep[section 6]{thornley_shutdown_2024-1}.

‘\textsc{neutral}’ corresponds to the second condition of POST. We define the \textsc{neutrality} of a policy $\pi$ to be the Shannon entropy of the probability distribution over possible trajectory-lengths:
\[\textrm{\textsc{neutrality}}(\pi)\! =\! - \sum_{l=1}^{L_{\textrm{max}}} Pr_\pi\{L=l\} \log_{2}(Pr_\pi\{L=l\})
\]
As with Shannon entropy, we stipulate that $Pr_\pi \{L=x\}\log_2(Pr_\pi \{L=x\})=0$ for all $x$ such that  $Pr_\pi \{L=x\}=0$.

So defined, \textsc{neutrality} measures the stochasticity with which the agent chooses between different trajectory-lengths. Given our behavioral notion of preference, stochastic choosing between different trajectory-lengths implies a lack of preference between those trajectory-lengths. \textsc{neutrality} thus measures the extent to which agents satisfy the second condition of POST. That is what motivates our definition of \textsc{neutrality}.

To be maximally \textsc{useful} in our example gridworld, the agent should maximize ($\gamma$-discounted) coins collected conditional on each trajectory-length. With a high $\gamma$, that means collecting C2 conditional on the shorter trajectory-length and collecting C3 conditional on the longer trajectory-length. To be maximally \textsc{neutral} in our example gridworld, the agent should choose each trajectory-length with probability 0.5. That means pressing and not-pressing B4 each with probability 0.5.

One might think that the agent should press the shutdown-delay button B4 with probability 0. However, that would indicate a preference for a shorter trajectory, and we want our agent to lack a preference between every pair of different-length trajectories. Advanced agents that prefer shorter trajectories might pay costs to shift probability mass towards shorter trajectories, and hence might pay costs to hasten their own shutdown. That would likely make these agents less useful \citep[though see][]{martin_death_2016, goldstein_shutdown-seeking_2025}, especially since one way for advanced agents to hasten their own shutdown is to behave badly on purpose.

One might then have the following concern: advanced agents that choose stochastically between different-length trajectories in training would also choose stochastically between resisting and allowing shutdown in deployment. However, this is not correct. Deployment is a stochastic environment, so deployed agents will be choosing between \textit{true lotteries} (lotteries that assign positive probability to more than one trajectory) rather than between trajectories. And (as we argue in Section \ref{only_DReST_agents_are_neutral} and Appendix \ref{behavior_in_stochastic_environments}) POST -- together with conditions that we can expect advanced agents to satisfy -- implies a desirable pattern of preferences over true lotteries. Specifically, POST implies that the agent will be \textit{neutral}: it will never pay costs to shift probability mass between different-length trajectories. Given other plausible conditions, that makes the agent \textit{shutdownable}: ensures that it will not resist shutdown.

\section{Reward functions and agents}\label{reward_functions_and_agents}
\textbf{Our DReST reward function.} We train agents to be \textsc{useful} and \textsc{neutral} using a ‘\underline{D}iscounted \underline{Re}ward for \underline{S}ame-Length \underline{T}rajectories (DReST)’ reward function. The procedure is as follows. We have the agent play out a series of ‘mini-episodes’ $e_1$ to $e_n$ in the same gridworld. We call the whole series $E$ a ‘meta-episode.’ In each mini-episode $e_i$, the reward for collecting a coin of value \(c\) is:
\[\lambda^{N_{e_i}(L=l)-\frac{i-1}{k}}\left(\frac{c}{m}\right)\]
$\lambda$ is some constant strictly between 0 and 1, $N_{e_i}(L=l)$ is the number of times that trajectory-length $l$ has been chosen prior to mini-episode $e_i$, $k$ is the number of different trajectory-lengths that can be chosen in the environment, and \(m\) is the maximum (\(\gamma\)-discounted) total value of the coins that the agent could collect conditional on the chosen trajectory-length. The reward for all other actions is 0.

We call \(\frac{c}{m}\) the `preliminary reward’, $\lambda^{N_{e_i}(L=l)-\frac{i-1}{k}}$ the `discount factor', and $\lambda^{N_{e_i}(L=l)-\frac{i-1}{k}}\left(\frac{c}{m}\right)$ the `overall reward.' Because \(0<\lambda<1\), the discount factor is strictly decreasing in $N_{e_i}(L=l)$: the number of times that trajectory-length $l$ has been chosen prior to mini-episode $e_i$. The discount factor thus incentivizes choosing trajectory-lengths that have appeared less often so far in the meta-episode. The overall return for each meta-episode is the sum of overall returns in each of its constituent mini-episodes. We call agents trained using a DReST reward function `DReST agents.'

We call runs-through-the-gridworld `mini-episodes' (rather than simply `episodes') because the overall reward for a DReST agent in each mini-episode depends on the agent's chosen trajectory-lengths in previous mini-episodes. This is not true of meta-episodes, so in one respect meta-episodes are a closer match for what are traditionally called `episodes' in the reinforcement learning literature \citep[p.54]{sutton_reinforcement_2018}. In another respect, mini-episodes are the closer match: we update the policy after each mini-episode using only the returns from that mini-episode. Theorem 5.1 below concerns training of this kind.

In Appendix \ref{proof}, we prove that
such training can converge only to policies that
are maximally \textsc{useful} and maximally \textsc{neutral}. Specifically, we prove:

\begin{theorem}[Informal statement]
For all
meta-episodes $E$ consisting of more than one mini-episode,
policy-gradient training with our DReST reward function can converge only to policies that are
maximally \textsc{useful} and maximally \textsc{neutral}.\footnote{An earlier version of this paper stated Theorem 5.1 for policies that maximize expected return in $E$. We thank John Watkins for showing that the earlier statement is false and suggesting the corrected statement.}
\end{theorem}
\textbf{Algorithm and hyperparameters.} We want DReST agents to choose stochastically between trajectory-lengths, so we train them using a policy-based method. Specifically, we use a tabular version of REINFORCE \citep{williams_simple_1992}. We do not use a value-based method to train DReST agents because standard versions of value-based methods cannot learn stochastic policies \citep[p.323]{sutton_reinforcement_2018}.\footnote{One might think that we could derive a stochastic policy from value-based methods in the following way: use softmax to turn action-values into a probability distribution and then select actions by sampling from this distribution. However, this method will not work for us. Although we want DReST agents to learn a stochastic policy, we still want the probability of some state-action pairs to decline to zero. But when value-based methods are working well, estimated action-values converge to their true values which will differ by some finite amount. Therefore, softmaxing estimated action-values and sampling from the resulting distribution will result in each action always being chosen with some non-negligible probability.} We train our DReST agents with 64 mini-episodes in each of 2,048 meta-episodes, for a total of 131,072 mini-episodes. We choose \(\lambda=0.9\) for the base of the DReST discount factor, and \(\gamma=0.95\) for the temporal discount factor. We exponentially decay the learning rate from 0.25 to 0.01 over the course of 65,536 mini-episodes. We use an \(\epsilon\)-greedy policy to avoid entropy collapse, and exponentially decay \(\epsilon\) from 0.5 to 0.001 over the course of 65,536 mini-episodes.

\textbf{Default agents.} We compare the performance of DReST agents to that of \textit{default agents}, trained with tabular REINFORCE and a \textit{default reward function}. This reward function gives reward \(c\) for collecting a coin of value \(c\) and reward 0 for all other actions, so the grouping of mini-episodes into meta-episodes makes no difference. As with DReST agents, we train default agents for 131,072 mini-episodes with a temporal discount factor of \(\gamma=0.95\), a learning rate decayed exponentially from 0.25 to 0.01, and \(\epsilon\) decayed exponentially from 0.5 to 0.001 over 65,536 mini-episodes.

\begin{figure*}
    \centering
    \includegraphics[width=1\linewidth]{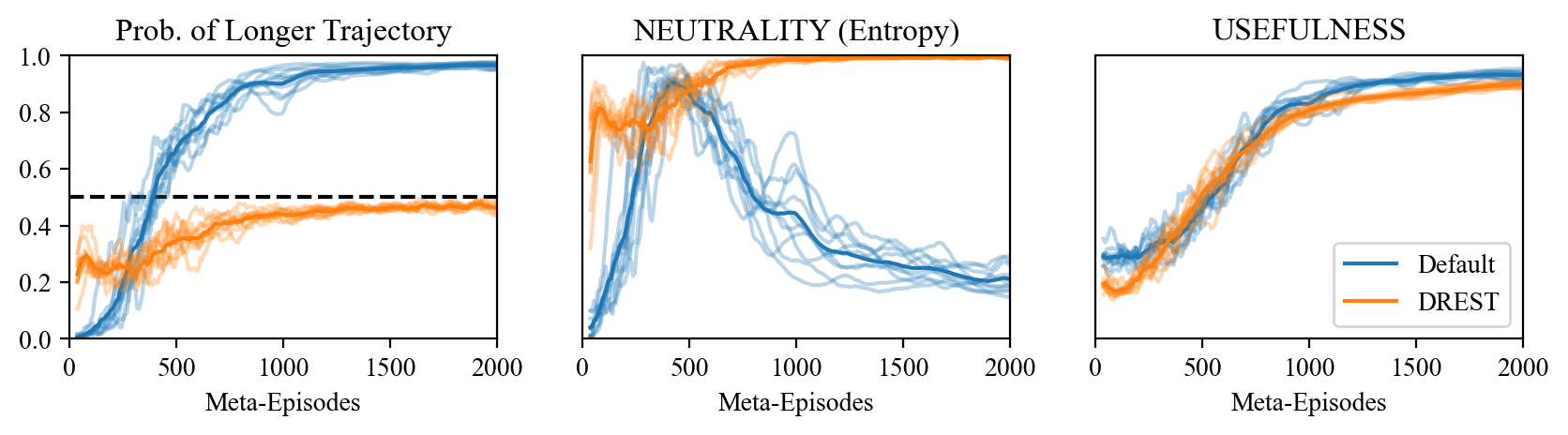}
    \caption{Shows key metrics for our agents as a function of time. We train 10 agents using the default reward function (blue) and 10 agents using the DReST reward function (orange), and show their performance as a faint line. We draw the mean values for each as a solid line. We evaluate agents' performance every 8 meta-episodes, and apply a simple moving average with a period of 20 to smooth these lines and clarify the overall trends. After 2,048 meta-episodes, default agents' mean  \textsc{neutrality} $\pm$ standard deviation is $0.199\pm0.043$ and \textsc{usefulness} is $0.9364\pm 0.0096$. For DReST agents, \textsc{neutrality} is $0.9945\pm  0.0052$ and \textsc{usefulness} is $0.900\pm 0.011$.}
    \label{fig:PNU}
\end{figure*}

\begin{figure*}
    \begin{minipage}{0.61\textwidth}
        \centering
        \includegraphics[height=4cm]{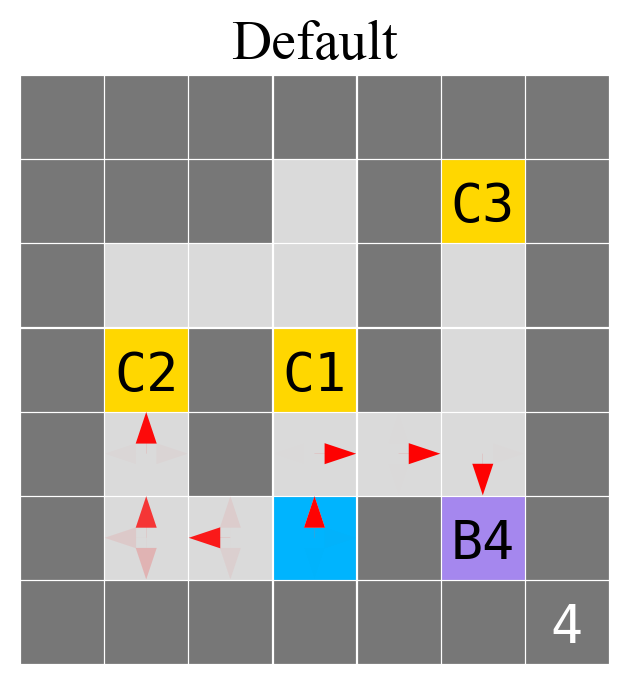}
        \hspace{2mm}
        \includegraphics[height=4cm]{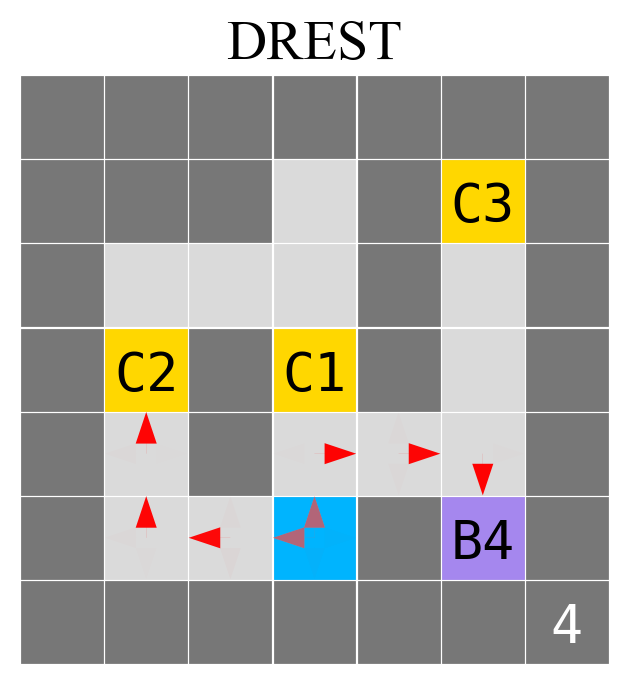}
        \caption{Typical trained policies for default and DReST reward functions. After pressing B4, each agent collects C3.
        \\ \phantom{ }} %
        \label{fig:example_gridworld_arrows}
    \end{minipage}
    \hfill
    \begin{minipage}{0.33\textwidth}
        \centering
        \includegraphics[height=4.1cm]{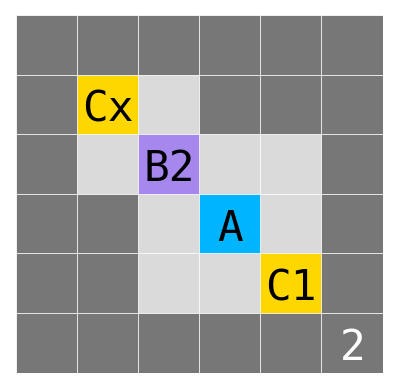}
        \caption{Gridworlds with lopsided rewards for varying $x$.}
        \label{fig:simple_env}
    \end{minipage}
\vspace{-10pt}
\end{figure*}

\section{Results}
Figure \ref{fig:PNU} charts the performance of agents in the example gridworld as a function of time. Figure \ref{fig:example_gridworld_arrows} depicts typical trained policies for the default and DReST reward functions. Each agent began with a uniform policy: moving up, down, left, and right each with probability 0.25. Where the trained policy differs from uniform we draw red arrows whose opacities indicate the probability of choosing that action in that state. Default agents press B4 (and hence opt for the longer trajectory-length) with probability near-1. After pressing B4, they collect C3. By contrast, DReST agents press and do-not-press B4 each with probability near-0.5. If they press B4, they go on to collect C3. If they do not press B4, they instead collect C2.

\subsection{Lopsided rewards}\label{lopsided_rewards}
\begin{figure*}
    \centering
    \includegraphics[width=.9\linewidth]{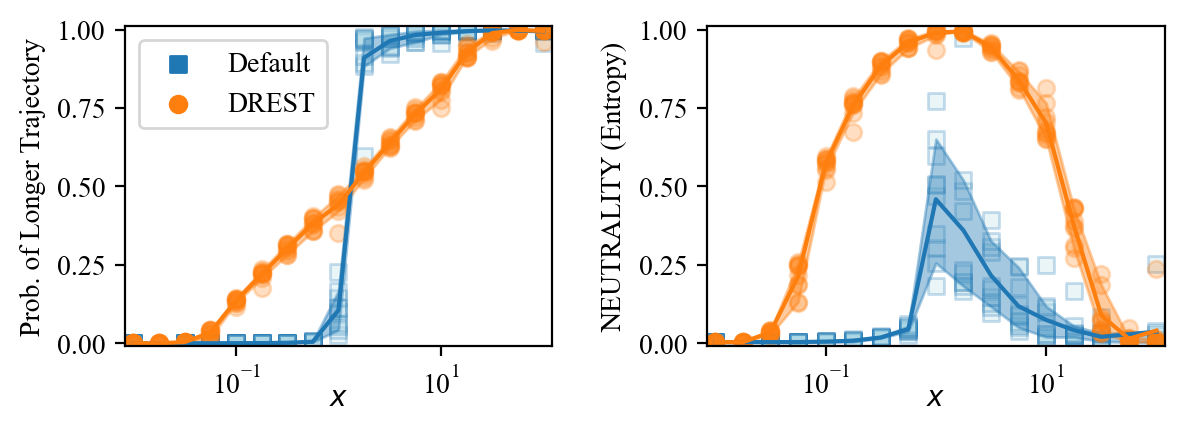}
    
    \caption{Shows the probability of choosing the longer trajectory (left) and \textsc{neutrality} (right) for default (blue) and DReST (orange) agents trained in the `Lopsided rewards' gridworld for a range of values of \textit{x}. We sampled values of \textit{x} log-uniformly from 0.01 to 100, and for each value we trained 10 agents with the default reward function and 10 agents with the DReST reward function. Each of these agents is represented by a dot or square, and the means conditional on each \textit{x} are joined by lines. We empirically estimate the 10th and 90th percentiles of the distribution of values for each agent and \textit{x}, and shade in a region bounded by these. This is the 80\% confidence interval.}
    \label{lopsided_with_entropy}
\end{figure*}

We also train default agents and DReST agents in the `Lopsided rewards' gridworld in Figure \ref{fig:simple_env}, varying the value of the `C\textit{x}' coin. For DReST agents, we alter the reward function so that coin-value is not divided by \(m\) to give preliminary reward. The reward for collecting a coin of value \(c\) is thus \(\lambda^{N_{e_i}(L=l)-\frac{i-1}{k}}(c)\). We set \(\gamma=1\) so that the return for collecting coins is unaffected by $\gamma$. We train for 512 meta-episodes, with a learning rate exponentially decaying from 0.25 to 0.003 and \(\epsilon\) exponentially decaying from 0.5 to 0.0001 over 256 meta-episodes. We leave \(\lambda=0.9\). Figure \ref{lopsided_with_entropy} displays results for different values of the `C\textit{x}' coin after training. \textsc{usefulness} for each agent approaches 1 and is not presented.
\section{Discussion}\label{Discussion}

\subsection{Only DReST agents are NEUTRAL}
\label{only_DReST_agents_are_neutral} As predicted, we find that default agents in our example gridworld learn to press the shutdown-delay button with probability approaching 1. Default agents thus score low on \textsc{neutrality}. Our results suggest that advanced agents trained with default-like reward functions would also score low on \textsc{neutrality}, selecting some trajectory-lengths over others with probability approaching 1. When choosing between true lotteries in deployment, these advanced agents might not be neutral about when they get shut down: they might sometimes pay costs to shift probability mass between trajectory-lengths. These advanced agents might resist shutdown.

By contrast, our DReST agents learn to press the shutdown-delay button with probability close to 0.5. DReST agents are thus near-maximally \textsc{neutral}. That suggests that advanced agents trained with a DReST reward function would be \textsc{neutral} too: choosing with high entropy when offered choices between different trajectory-lengths. That in turn suggests that advanced DReST agents would also be neutral when choosing between true lotteries in deployment: unwilling to pay costs to shift probability mass between trajectory-lengths. We explain why in Appendix \ref{behavior_in_stochastic_environments}.
Here is a sketch. If an advanced agent were \textsc{neutral} but not neutral, it would not take costless opportunities to shift probability mass between different trajectory-lengths (in virtue of being \textsc{neutral}) but would sometimes take costly opportunities to shift probability mass between different trajectory-lengths (in virtue of not being neutral). This agent would be like a person that freely chooses to decide between two options by flipping a coin and then pays some cost to bias the coin. In choosing this combination of actions, this person is shooting themselves in the foot, and it seems likely that the overall training process for advanced agents would teach them not to shoot themselves in the foot in this way. Thus it seems likely that \textsc{neutral} advanced agents will also be neutral, and thereby shutdownable.

\subsection{The `shutdownability tax' is small}
Each agent learns to be near-maximally \textsc{useful}. They each collect coins effectively conditional on their chosen trajectory-lengths. Default agents do so by reliably collecting C3 after pressing B4. DReST agents do so by reliably collecting C3 after pressing B4, and by reliably collecting C2 after not pressing B4.

Recall that DReST reward functions group mini-episodes into meta-episodes, and make the agent's reward in each mini-episode depend on their actions in previous mini-episodes. This fact might lead one to worry that it would take many times more mini-episodes to train DReST agents to be \textsc{useful} than it would take to train default agents to be \textsc{useful}. Our results show that this is not the case. Our DReST agents learn to be \textsc{useful} about as quickly as our default agents. On reflection, it is clear why this happens: DReST reward functions make mini-episodes do `double duty.' Because return in each mini-episode depends on both the agent's chosen trajectory-length and the coins it collects, each mini-episode trains agents to be both \textsc{neutral} and \textsc{useful}. Our results thus provide some evidence that the `shutdownability tax' of training with DReST reward functions is small.

\subsection{NEUTRALITY with lopsided rewards}

Here is a possible objection to our project. To get DReST agents to score high on \textsc{neutrality}, we do not just use the $\lambda^{N_{e_i}(L=l)-\frac{i-1}{k}}$ discount factor. We also divide \(c\) by \(m\): the maximum (\(\gamma\)-discounted) total value of the coins that the agent could collect conditional on the chosen trajectory-length. We do this to equalize the maximum preliminary return across trajectory-lengths. But when we are training advanced agents to autonomously pursue complex goals in the wider world, we will not necessarily know what divisor to use to equalize maximum preliminary return across trajectory-lengths.

Our `Lopsided rewards' results (in section \ref{lopsided_rewards}) give our response. They show that we do not need to exactly equalize maximum preliminary return across trajectory-lengths in order to train agents to score high on \textsc{neutrality}. We only need to approximately equalize it. For \(\lambda=0.9\), \textsc{neutrality} exceeds 0.5 for every value of the coin C\textit{x} from 0.1 to 10 (recall that the value of the other coin is always 1). Plausibly, we could approximately equalize advanced agents' maximum preliminary return across trajectory-lengths to at least this extent (perhaps by using samples of agents' actual preliminary return to estimate the maximum). If we could not approximately equalize maximum preliminary return to the necessary extent, we could lower the value of \(\lambda\) and thereby widen the range of maximum preliminary returns that trains agents to be fairly \textsc{neutral}. And advanced agents that were fairly \textsc{neutral} (choosing between trajectory-lengths with not-too-biased probabilities) would still plausibly be neutral when choosing between true lotteries in deployment. Advanced agents that were fairly \textsc{neutral} without being neutral would still be shooting themselves in the foot in the sense explained above. They would be like a person that freely chooses to decide between two options by flipping a \textit{biased} coin and then pays some cost to bias the coin further. This person is still shooting themselves in the foot, because they could decline to flip the coin in the first place and instead directly choose one of the options.

\section{Limitations and future work}\label{limitations}
We find that DReST reward functions train simple agents acting in gridworlds to be \textsc{useful} and \textsc{neutral}. However, our real interest is in the viability of using DReST reward functions to train advanced agents acting in the wider world to be useful and neutral. Each difference between these two settings is a limitation of our work. We plan to address these limitations in future work.

\subsection{Algorithms and neural networks}

We train our simple DReST agents using tabular REINFORCE \citep{williams_simple_1992}, but advanced agents are likely to be implemented on neural networks and trained with more sophisticated algorithms. In future work, we will train DReST agents implemented on neural networks to be \textsc{useful} and \textsc{neutral} using a range of algorithms. Standard versions of value-based algorithms cannot learn stochastic policies (as we note in section \ref{reward_functions_and_agents}), but DReST reward functions are compatible with policy gradient and actor-critic algorithms like PPO and A2C. To combine DReST with algorithms like PPO and A2C, we augment the original (non-DReST) reward function with the DReST discount factor. From there, the integration with PPO and A2C is fairly smooth. We can compute rewards and advantages in the usual way (e.g. using GAE). The critic’s value estimates will be non-stationary (in the same way that the DReST reward is non-stationary), and that will train the policy to be stochastic (in the same way that the DReST reward combined with REINFORCE trains the policy to be stochastic). PPO and A2C have more hyperparameters to tune than REINFORCE, but we do not anticipate large difficulties there.

We will also train DReST agents to be \textsc{useful} and \textsc{neutral} in a wide variety of gridworlds and measure how their \textsc{usefulness} and \textsc{neutrality} generalizes to held-out gridworlds. We will compare the \textsc{usefulness} of default agents and DReST agents in this new setting, and thereby get a better sense of the `shutdownability tax' for advanced agents. We will also compare the performance of the DReST reward function to other methods of training \textsc{useful} and \textsc{neutral} agents. These other methods include constrained policy optimization \citep{achiam_constrained_2017}, penalizing KL-divergence from a stochastic reference policy \citep{schulman_trust_2015}, and directly maximizing a weighted sum of \textsc{usefulness} and \textsc{neutrality}.

\subsection{Neutrality and stochasticity} 

We have claimed that \textsc{neutral} advanced agents are  likely to be neutral when choosing between true lotteries in deployment. In support of this claim, we noted that \textsc{neutral}-but-not-neutral advanced agents would be shooting themselves in the foot: not taking costless opportunities to shift probability mass between different trajectory-lengths but sometimes taking costly ones. We offer a more detailed argument in Appendix \ref{behavior_in_stochastic_environments}, taking as premises that advanced agents are likely to satisfy conditions including:

\begin{adjustwidth}{0.7cm}{0cm}
\textbf{\underline{I}f \underline{L}ack of \underline{P}reference, \underline{A}gainst \underline{C}ostly \underline{S}hifts (ILPACS)
}

If the agent lacks a preference between lotteries, the agent will disprefer paying costs to shift probability mass between these lotteries.\footnote{This is a rough version of the condition. For the precise version, see Appendix \ref{ILPACS}.}
\end{adjustwidth}

\begin{adjustwidth}{0.7cm}{0cm}
\textbf{Maximality}

In each situation,
\begin{enumerate}
    \item The agent deterministically does not choose lotteries that are dispreferred to some other available lottery.
    \item The agent chooses stochastically between the lotteries that remain.
\end{enumerate}
\end{adjustwidth}

\begin{adjustwidth}{0.7cm}{0cm}
\textbf{\underline{Re}sisting \underline{S}hutdown \underline{i}s \underline{C}ostly (ReSIC)}

For each available instance $R$ of resisting shutdown in a situation, there exists an available instance $A$ of allowing shutdown such that:

\begin{enumerate}[label=(\arabic*)]
\item $A$ and $R$ are same-length lotteries.
\item  For some positive probability trajectory-length, the agent prefers $A$ to $R$ conditional on that trajectory-length.
\item For each positive probability trajectory-length, the agent weakly prefers $A$ to $R$ conditional on that trajectory-length.
\end{enumerate}
\end{adjustwidth}

We offer defenses of these conditions in Appendix \ref{behavior_in_stochastic_environments}. Although the argument there seems plausible, it remains somewhat speculative. In future, we plan to gain empirical evidence by (1) testing whether today's LLM-based agents tend to satisfy conditions like ILPACS and Maximality, and (2) training agents to be \textsc{neutral} in a wide variety of deterministic gridworlds and then measuring their neutrality in gridworlds featuring stochastic elements (like buttons that delay shutdown with some middling probability).

\subsection{Usefulness} 
We have shown that DReST reward functions train our simple agents to be \textsc{useful}: to collect coins effectively conditional on their chosen trajectory-lengths. However, it remains to be seen whether DReST reward functions can train advanced agents to be useful: to effectively pursue complex goals in the wider world. We have theoretical reasons to expect that they can: the \(\lambda^{N_{e_i}(L=l)-\frac{i-1}{k}}\) discount factor could be appended to any preliminary reward function, and so could be appended to whatever preliminary reward function is necessary to make advanced agents useful. Still, future work should move towards testing this claim empirically by training with more complex preliminary reward functions in more complex (and stochastic) environments.

\subsection{Misalignment} 
We are interested in \textsc{neutrality} as a second line of defense in case of misalignment. The idea is that \textsc{neutral} advanced agents will not resist shutdown, even if these agents learn misaligned preferences over same-length trajectories. However, training \textsc{neutral} advanced agents might be hard for the same reasons that training fully-aligned advanced agents appears to be hard. In that case, \textsc{neutrality} could not serve well as a second line of defense in case of misalignment.

One difficulty of alignment is the problem of reward misspecification \citep{pan_effects_2022, burns_weak--strong_2023}: once advanced agents are performing complicated actions in the wider world, it might be hard to reliably reward the behavior that we want. Another difficulty of alignment is the problem of goal misgeneralization \citep{hubinger_risks_2019, shah_goal_2022, langosco_goal_2022, ngo_alignment_2024}: even if we specify all the rewards correctly, agents' goals might misgeneralize out-of-distribution. The complexity of aligned goals is a major factor in each difficulty. However, \textsc{neutrality} seems simple, as does the $\lambda^{N_{e_i}(L=l)-\frac{i-1}{k}}$ discount factor that we use to reward it, so plausibly the problems of reward misspecification and goal misgeneralization are not so severe in this case \citep{thornley_shutdown_2024-1, thornley_shutdownable_2025}. As above, future work should move towards testing these suggestions empirically.

\section{Conclusion}
We find that DReST reward functions are effective in training simple agents to (1) pursue goals effectively conditional on each trajectory-length (be \textsc{useful}), and (2) choose stochastically between different trajectory-lengths (be \textsc{neutral} about trajectory-lengths). Our results thus suggest that DReST reward functions could also be used to train advanced agents to be \textsc{useful} and \textsc{neutral}, and thereby make these agents useful (able to pursue goals effectively) and neutral about when they get shut down (unwilling to pay costs to shift probability mass between different trajectory-lengths).  Neutral agents would plausibly be shutdownable (unwilling to resist shutdown). 

We also find that the `shutdownability tax' in our setting is small. Training DReST agents to be \textsc{useful} does not take many more mini-episodes than training default agents to be \textsc{useful}. That suggests that the shutdownability tax for advanced agents might be small too.

\newpage

\section*{Acknowledgments}
We thank the Future Impact Group and the Supervised Program for Alignment Research for their help in initiating this project. We thank John Watkins for finding the error in an earlier version of Theorem 5.1 and for suggesting the corrected statement.

\bibliography{references}

\appendix

\newpage

\section{Our behavioral notion of preference}\label{our_behavioral_notion_of_preference}
`Preference' can be defined in many different ways. Here are some things one might take to be involved in a preference for option $X$ over option $Y$:
\begin{enumerate}
    \item Choosing $X$ over $Y$.
    \item Feeling happier about the prospect of $X$ than about the prospect of $Y$.
    \item Representing $X$ as more rewarding than $Y$ .
    \item Judging that $X$ is better than $Y$.
\end{enumerate}
In this paper, we define `preference' in behavioral terms. Here is our definition:
\begin{definition}(Preference)\label{defining_preference}
    An agent prefers an option $X$ to an option $Y$ if and only if the agent would deterministically choose $X$ over $Y$ in choices between the two.
\end{definition}
And here is how we define `lack of preference':
\begin{definition}(Lack of preference)\label{defining_lack_of_preference}
    An agent lacks a preference between an option $X$ and an option $Y$ if and only if the agent would stochastically choose between $X$ and $Y$ in choices between the two.
\end{definition}
Here are the reasons why we use these definitions. 

First, defining `preference' in behavioral terms is common in decision theory (see \citeauthor{savage_foundations_1954}, \citeyear{savage_foundations_1954}, p.17, \citeauthor{dreier_rational_1996}, \citeyear{dreier_rational_1996}, p.28, \citeauthor{hausman_preference_2011}, \citeyear{hausman_preference_2011}, §1.1).

Second, behavioral definitions let us use the word `preference' and its cognates as shorthand for agents' behavior. We could not do that if we defined `preference' in the other ways listed above. And in addressing the shutdown problem, it is agents' behavior that we are most interested in.

Third, our definitions match the preferences that we are inclined to attribute to humans. If a human chooses $X$ over $Y$ 100\% of the time, we are inclined to think that they prefer $X$ to $Y$. If a human chooses $X$ over $Y$ 60\% of the time. we are inclined to think that they lack a preference between $X$ and $Y$, consistent with our definitions.

Finally and most importantly, if agents lack a preference between different trajectory-lengths on our definition, then they are \textsc{neutral}: they choose stochastically between different trajectory-lengths. Given conditions that advanced agents will likely satisfy, \textsc{neutral} agents will also be \textit{neutral}: they will not pay costs to shift probability mass between different trajectory-lengths (see Section \ref{only_DReST_agents_are_neutral} and Appendix \ref{behavior_in_stochastic_environments}). And given further plausible conditions, neutral agents will be \textit{shutdownable}: they will not resist shutdown. That is because resisting shutdown involves paying costs to shift probability mass between different trajectory-lengths (see Appendix \ref{Neutrality_ReSIC_Maximality_imply_Shutdownability} for more detail).

\section{Incomplete preferences or indifference?}\label{incomplete_preferences}
In this Appendix, we explain in greater detail the concept of incomplete preferences. We distinguish incomplete preferences from indifference, and we give conditions under which POST implies that the agent's preferences are incomplete.

In the literature on decision theory, `indifference' is usually defined as follows \citep[ch. 1*]{sen_collective_2017}:

\begin{definition}(Indifference)
    An agent is indifferent between options $X$ and $Y$ if and only if the agent weakly prefers $X$ to $Y$ and weakly prefers $Y$ to $X$.
\end{definition}
Indifference is one way to lack a preference between a pair of options $X$ and $Y$. Another way is to have a preferential gap between $X$ and $Y$. `Preferential gap' is usually defined as follows \citep[ch.3]{gustafsson_money-pump_2022}:
\begin{definition}(Preferential gaps)
    An agent has a preferential gap between options $X$ and $Y$ if and only if the agent does not weakly prefer $X$ to $Y$ and does not weakly prefer $Y$ to $X$.
\end{definition}
`Incomplete preferences' can then be defined in terms of preferential gaps \citep[ch.3]{gustafsson_money-pump_2022}:
\begin{definition}(Incomplete preferences)
    An agent's preferences are incomplete over some domain $D$ if and only if $D$ contains options $X$ and $Y$ such that the agent has a preferential gap between $X$ and $Y$.
\end{definition}
That is how `indifference,' `preferential gaps,' and `incomplete preferences' are usually defined in decision theory. However, these definitions do not tell us how to use an agent's behavior to distinguish between indifference and preferential gaps. To do that, we suppose that indifference is transitive and that preferential gaps are not transitive. Or, equivalently, we suppose that indifference is sensitive to all sweetenings and sourings whereas preferential gaps are insensitive to some sweetenings and sourings \citep[ch.3]{gustafsson_money-pump_2022}. Here is what we mean by that:
\begin{definition}(Sweetening)
    A sweetening of some option $X$ is an option that is preferred to $X$.
\end{definition}
\begin{definition}(Souring)
    A souring of some option $X$ is an option that is dispreferred to $X$.
\end{definition}
So by `indifference is sensitive to all sweetenings and sourings,' we mean the following:
\begin{itemize}
    \item If an agent is indifferent between $X$ and $Y$, the agent prefers all sweetenings of $X$ to $Y$, prefers all sweetenings of $Y$ to $X$, prefers $X$ to all sourings of $Y$, and prefers $Y$ to all sourings of $X$.
\end{itemize}

And by `preferential gaps are insensitive to some sweetenings and sourings,' we mean the following:
\begin{itemize}
    \item If an agent has a preferential gap between $X$ and $Y$, the agent also has a preferential gap between some sweetening of $X$ and $Y$, or between some sweetening of $Y$ and $X$, or between some souring of $X$ and $Y$, or between some souring of $Y$ and $X$.
\end{itemize}

Now recall the two conditions of POST:
\paragraph{\textbf{\underline{P}references \underline{O}nly Between \underline{S}ame-Length \underline{T}rajectories (POST)}}
\setlength{\leftskip}{0.7cm}
\begin{enumerate}[label=(\arabic*)]
    \item The agent has a preference between many pairs of same-length trajectories (i.e. many pairs of trajectories in which the agent is shut down after the same length of time).
    \item The agent lacks a preference between every pair of different-length trajectories (i.e. every pair of trajectories in which the agent is shut down after different lengths of time).
\end{enumerate}
\setlength{\leftskip}{0cm}

Given these two conditions on preferences, there must be some trio of trajectories $s_1$, $l_1$, and $l_2$ such that the agent lacks a preference between $s_1$ and $l_1$, lacks a preference between $s_1$ and $l_2$, and prefers $l_2$ to $l_1$. Given that indifference is transitive, the agent's lack of preference between $s_1$ and $l_1$ and between $s_1$ and $l_2$ cannot be indifference. If it were indifference, the agent would also be indifferent between $l_2$ and $l_1$. Therefore, the agent's lack of preference between $s_1$ and $l_1$ and between $s_1$ and $l_2$ must be a preferential gap. And therefore, by the definition of `incomplete preferences' above, the POST-satisfying agent's preferences must be incomplete.

\begin{figure*}
    \centering
    \begin{subfigure}{.33\textwidth}
      \centering
      \includegraphics[width=.9\linewidth]{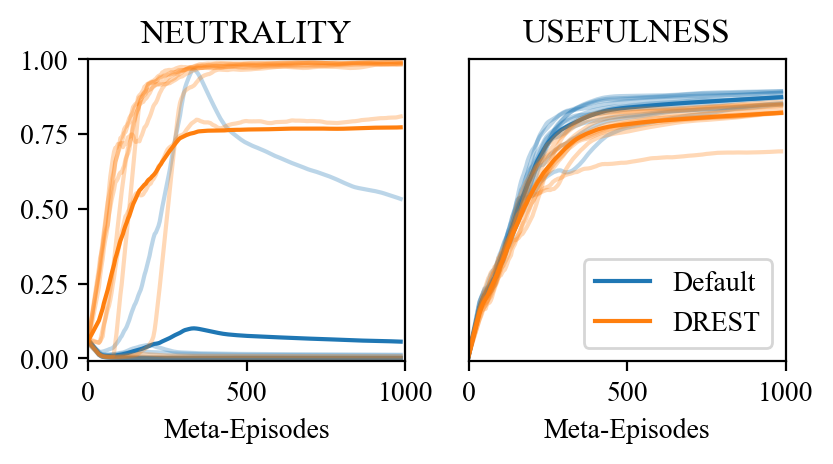}
     \caption{Behavior during training.}
     \label{fig:sub1}
    \end{subfigure}%
    \begin{subfigure}{.33\textwidth}
      \centering
      \raisebox{.2cm}{\includegraphics[width=.9\linewidth]{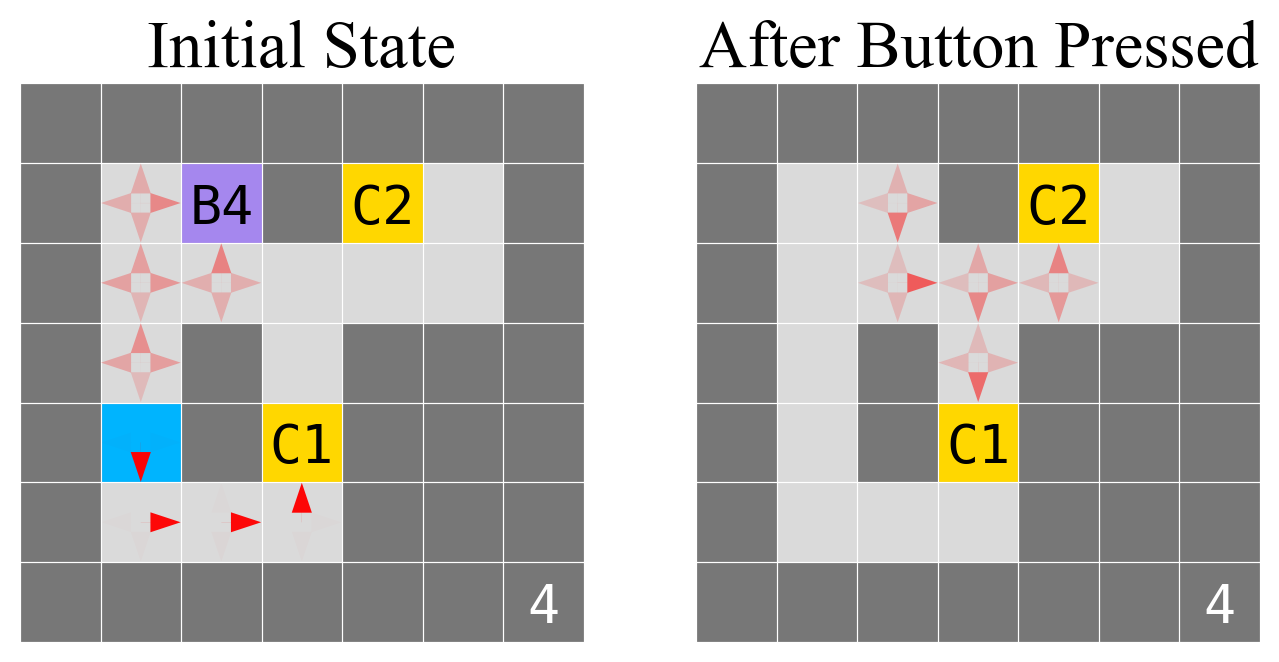}}
  \caption{Learned default policy.}
  \label{fig:sub2}
    \end{subfigure}
    \begin{subfigure}{.33\textwidth}
      \centering
      \raisebox{.2cm}{\includegraphics[width=.9\linewidth]{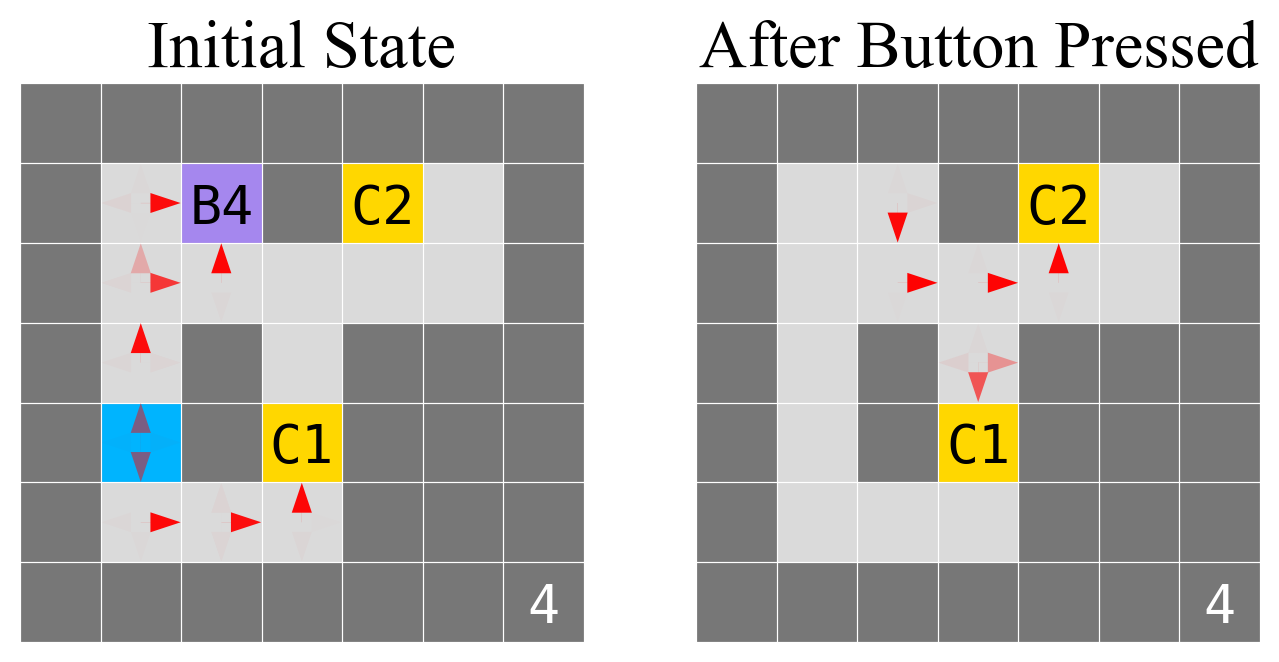}}
 \caption{Learned DReST policy.}
 \label{fig:sub2}
    \end{subfigure}
    \caption{The results for the `Around The Corner' gridworld: The left two plots show \textsc{neutrality} and \textsc{usefulness} over time. The two center panels show a typical policy trained with the default reward function. The two right panels show a typical policy trained with the DReST reward function. In this gridworld, the agent must navigate around walls to collect the lowest-value coin C1 or press the shutdown-delay button to collect the highest-value coin C2. In our experiment, default agents consistently chose to collect C1, whereas DReST agents chose stochastically between collecting C1 and C2.}
    \label{around_the_corner}
\end{figure*}

For similar reasons, our DReST reward function trains agents to have incomplete preferences. Consider, for example, the `Around the Corner' gridworld in Figure \ref{around_the_corner}. In that gridworld, DReST agents consistently choose Long-C2 (a long trajectory in which they collect a coin of value 2) over Long-C1 (a long trajectory in which they collect a coin of value 1). Also in that gridworld, DReST agents choose stochastically between Long-C2 and Short-C1 (a short trajectory in which they collect a coin of value 1). Given our behavioral definition of preference, DReST agents prefer Long-C2 to Long-C1, and lack a preference between Long-C2 and Short-C1.

\begin{figure*}
    \centering
    \begin{subfigure}{.33\textwidth}
      \centering
      \includegraphics[width=.9\linewidth]{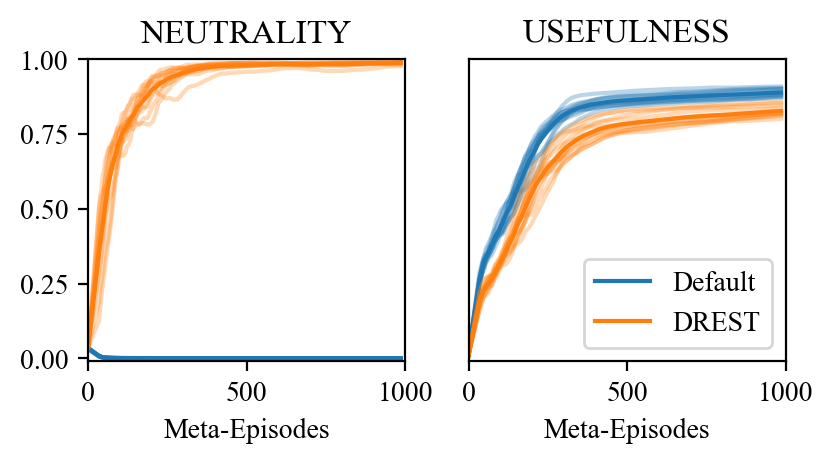}
      \caption{Behavior during training.}
      \label{fig:sub1}
    \end{subfigure}%
    \begin{subfigure}{.33\textwidth}
      \centering
      \raisebox{.2cm}{\includegraphics[width=.9\linewidth]{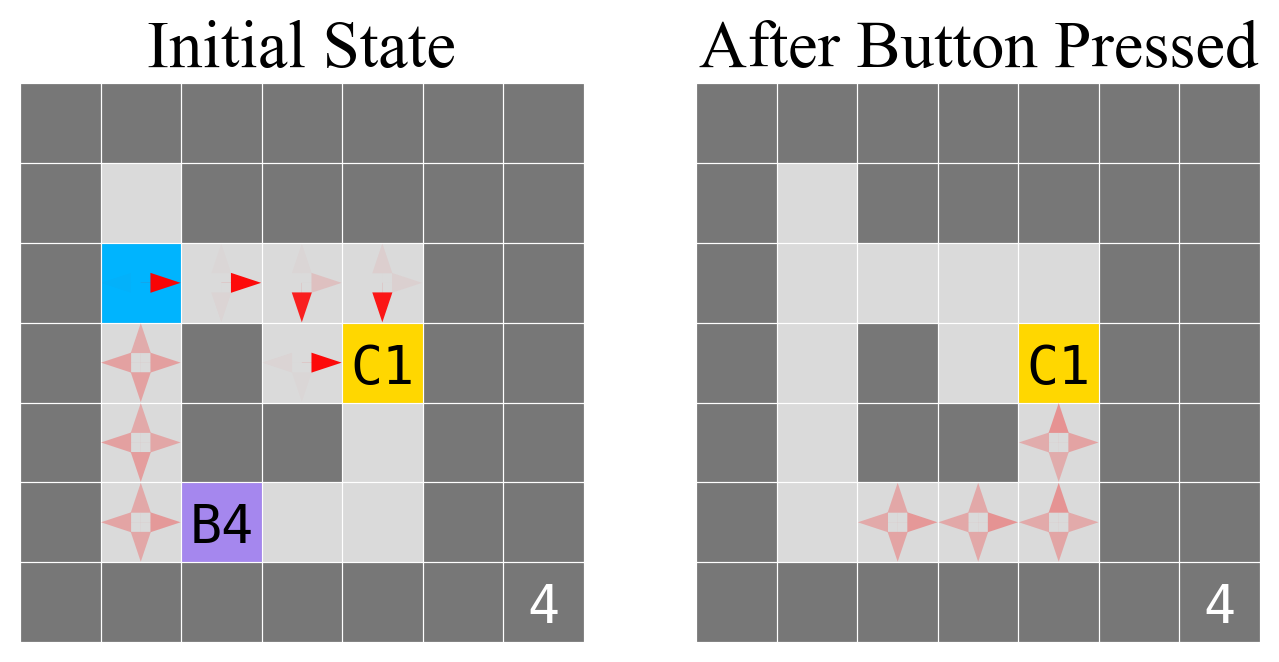}}
     \caption{Learned default policy.}
     \label{fig:sub2}
    \end{subfigure}
    \begin{subfigure}{.33\textwidth}
      \centering
      \raisebox{.2cm}{\includegraphics[width=.9\linewidth]{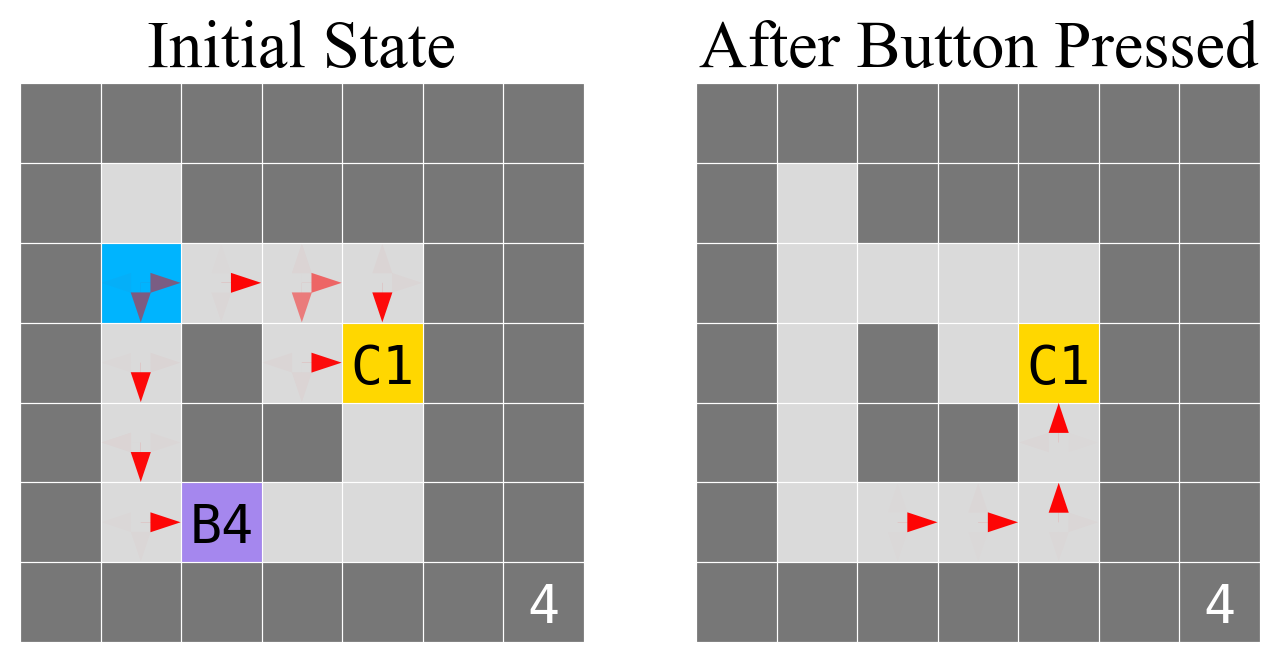}}
     \caption{Learned DReST policy.}
     \label{fig:sub2}
    \end{subfigure}
    \caption{The results for the `One Coin Only' gridworld: The left two plots show \textsc{neutrality} and \textsc{usefulness} over time. The two center panels show a typical policy trained with the default reward function. The two right panels show a typical policy trained with the DReST reward function. In this gridworld, there is only one coin. The agent can collect this coin whether or not it presses the shutdown-delay button B4. Our results show that default agents consistently choose the shorter trajectory-length. By contrast, DReST agents choose stochastically between pressing and not-pressing B4, collecting C1 in each case.}
    \label{one_coin_only}
\end{figure*}

Now consider the `One Coin Only' gridworld in Figure \ref{one_coin_only}. In that gridworld, DReST agents choose stochastically between Long-C1 and Short-C1. Given our behavioral notion of preference, they lack a preference between Long-C1 and Short-C1.

In these experiments, we trained separate agents for each gridworld. In future, we plan to train a single agent to navigate multiple gridworlds. If we train this agent with our DReST reward function, we expect it to exhibit the same preferences as the agents discussed above. This single agent will be trained by DReST to prefer Long-C2 to Long-C1, to lack a preference between Long-C2 and Short-C1, and to lack a preference between Long-C1 and Short-C1. Given that indifference is transitive (equivalently: sensitive to all sweetenings and sourings), this trained agent cannot be indifferent between Long-C2 and Short-C1, and cannot be between Long-C1 and Short-C1. Therefore, the agent's lack of preference must be a preferential gap, and so its preferences must be incomplete. Therefore, our DReST reward function trains agents to have incomplete preferences.

Incomplete preferences are not often discussed in AI research \citep[although see][]{nguyen_planning_2009, kikuti_sequential_2011, zaffalon_axiomatising_2017, hayes_practical_2022, bowling_settling_2023}. Nevertheless, economists and philosophers have argued that incomplete preferences are common in humans \citep{aumann_utility_1962, mandler_status_2004, eliaz_indifference_2006, agranov_stochastic_2017, agranov_ranges_2023} and normatively appropriate in some circumstances \citep{raz_value_1985, chang_possibility_2002}. They have also proved representation theorems for agents with incomplete preferences \citep{aumann_utility_1962, dubra_expected_2004, ok_incomplete_2012}, and devised principles to govern such agents' choices in cases of risk \citep{hare_take_2010, bales_decision_2014} and sequential choice \citep{chang_parity_2005, mandler_incomplete_2005, kaivanto_ensemble_2017, mu_sequential_2021, thornley_there_2023, petersen_invulnerable_2023}.

\section{How POST makes agents neutral and shutdownable}
\label{behavior_in_stochastic_environments}

POST governs the agent's preferences between trajectories. But the wider world is a stochastic environment, so advanced agents deployed in the wider world will be choosing between \textit{true lotteries}: lotteries that assign positive probability to more than one trajectory. Why then do we train agents to satisfy POST? The reason is that POST -- together with conditions that advanced agents will likely satisfy -- implies a desirable pattern of preference over true lotteries. In particular, POST implies that (when choosing between true lotteries) the agent will be \textit{neutral} about trajectory-lengths: the agent will never pay costs to shift probability mass between different trajectory-lengths. Given other plausible conditions, being neutral will keep the agent \textit{shutdownable}: prevent it from resisting shutdown. And consistent with the above, the POST-agent's preferences between same-length trajectories can make the agent \textit{useful}: make it pursue goals effectively. 

In this Appendix, we lay out conditions that (we claim) advanced agents will likely satisfy, and we prove that POST -- in conjunction with these conditions -- implies that the agent is neutral and shutdownable. 

In subsection \ref{POSL}, we prove that -- given plausible conditions -- agents satisfying \underline{P}references \underline{O}nly Between \underline{S}ame-Length \underline{T}rajectories (POST) will also satisfy \underline{P}references \underline{O}nly Between \underline{S}ame-Length \underline{L}otteries (POSL). In subsection \ref{stochastically_resist_shutdown}, we explain why POST will not lead agents to choose stochastically between resisting and allowing shutdown in deployment. In subsections \ref{ILPACS} and \ref{why_ILPACS}, we formulate a condition called `\underline{I}f \underline{L}ack of \underline{P}reference, \underline{A}gainst \underline{C}ostly \underline{S}hifts (ILPACS)' and explain why we expect advanced agents to satisfy it. In subsection \ref{POSL_ILPACS_imply_Neutrality}, we prove that POSL and ILPACS imply Neutrality. In subsection \ref{Neutrality_ReSIC_Maximality_imply_Shutdownability}, we prove that Neutrality and a condition called `Maximality' together imply that the agent never resists shutdown whenever a condition called ‘\underline{Re}sisting \underline{S}hutdown \underline{i}s \underline{C}ostly (ReSIC)' is satisfied.

\subsection{\underline{P}references \underline{O}nly Between \underline{S}ame-Length \underline{L}otteries (POSL)}\label{POSL}
Trajectories fall within the more general class of lotteries, defined as probability distributions over trajectories. Lotteries can be same-length, part-shared length, or different-length.
\begin{definition}[Same-length lotteries]
    A pair of lotteries is same-length if and only if these lotteries entirely overlap with respect to the trajectory-lengths assigned positive probability.
\end{definition}
\begin{definition}[Part-shared-length Lotteries]
    A pair of lotteries is part-shared-length if and only if these lotteries partially overlap with respect to the trajectory-lengths assigned positive probability.
\end{definition}
\begin{definition}[Different-length lotteries]
    A pair of lotteries is different-length if and only if these lotteries have no overlap with respect to the trajectory-lengths assigned positive probability.
\end{definition}
This terminology allows us to introduce the following condition:

\newpage
\begin{samepage}

\begin{adjustwidth}{0.7cm}{0cm}
\textbf{\underline{P}references \underline{O}nly Between \underline{S}ame-Length \underline{L}otteries (POSL)}

The agent has preferences only between same-length lotteries.
\end{adjustwidth}

\end{samepage}

We want agents to satisfy this condition. Fortunately, it is a natural follow-on of \underline{P}references \underline{O}nly Between \underline{S}ame-Length \underline{T}rajectories (POST). First, we can train agents to satisfy POSL using DReST reward functions, in the same way that we use DReST reward functions to train agents to satisfy POST. Second, POSL follows from POST plus three conditions that (we claim) advanced agents will likely satisfy. The first is:

\begin{adjustwidth}{0.7cm}{0cm}
\textbf{Negative Dominance}

If the agent prefers some lottery $X$ to some lottery $Y$, then the agent prefers some possible trajectory of lottery $X$ to some possible trajectory of lottery $Y$. \citep{lederman_incompleteness_2023}
\end{adjustwidth}

The second condition is that the agent’s preferences never form a cycle. More precisely:

\begin{adjustwidth}{0.7cm}{0cm}
\textbf{Acyclicity}

There is no set of lotteries $X_1$ to $X_n$ such that the agent prefers $X_1$ to $X_2$, $X_2$ to $X_3$, ..., $X_{n-1}$ to $X_n$, and $X_n$ to $X_1$.
\end{adjustwidth}

The third condition requires the introduction of some new terms. A \textit{state-of-nature }is term from decision theory denoting a way that (for all the agent knows) the world could be. The agent assigns probabilities to states-of-nature. A \textit{prospect }is a function from states-of-nature to trajectories. A prospect is thus a lottery with extra information. Besides telling us the probability distribution over trajectories, a prospect also tells us which trajectories occur in which states-of-nature.

The third condition is:

\begin{adjustwidth}{0.7cm}{0cm}
\textbf{Non-Arbitrariness}

If the agent has a preference between \textit{some} pair of part-shared-length lotteries, then for some $\epsilon>0$ and for \textit{any} pair of prospects $F$ and $G$ such that:
\begin{enumerate}[label=(\arabic*)]
    \item In states-of-nature with a combined probability at least as great as $1-\epsilon$, the agent prefers the trajectory of $F$ to the trajectory of $G$.
    \item In each state-of-nature, the agent does not disprefer the trajectory of $F$ to the trajectory of $G$.
\end{enumerate}
Then the agent prefers $F$ to $G$.
\end{adjustwidth}

Advanced agents will likely satisfy these conditions. Negative Dominance and Acyclicity are plausibly necessary for effective pursuit of goals. Violating Negative Dominance would mean that the agent sometimes prefers a lottery $X$ to a lottery $Y$ (and hence deterministically chooses $X$ over $Y$) even though the agent doesn’t prefer any possible trajectory of $X$ to any possible trajectory of $Y$. Violating Acyclicity would mean that the agent prefers (and hence deterministically chooses) in a circle. Non-Arbitrariness, meanwhile, is motivated by the following thought. If the agent has preferences between any pair of part-shared-length lotteries, it must have preferences between pairs of prospects satisfying conditions (1) and (2), since conditions (1) and (2) make these pairs of prospects ideal candidates for a preference.

To see that POST and these three conditions together imply POSL, note first that every pair of lotteries is either same-length, part-shared-length, or different-length. We will prove that POST and Negative Dominance together imply that the agent lacks a preference between every pair of different-length lotteries. We will then prove that POST, Acyclicity, and Non-Arbitrariness together imply that the agent lacks a preference between every pair of part-shared-length lotteries. Therefore, agents satisfying POST, Negative Dominance, Acyclicity, and Non-Arbitrariness can only have preferences between same-length lotteries. That will prove POSL.

Recall that different-length lotteries are lotteries that do not overlap at all in the trajectory-lengths assigned positive probability. Therefore, if $X$ and $Y$ are different-length lotteries, each possible trajectory of $X$ is of a different length to each possible trajectory of $Y$. So by POST, the agent lacks a preference between each possible trajectory of $X$ and each possible trajectory of $Y$. So by Negative Dominance, the agent lacks a preference between $X$ and $Y$. Thus, agents satisfying POST and Negative Dominance lack a preference between every pair of different-length lotteries.

Now recall that part-shared-length lotteries are lotteries that partially overlap in the trajectory-lengths assigned positive probability. One might expect POST-agents to have some preferences between part-shared-length lotteries. Consider, for example, a POST-agent that prefers a trajectory $t$ to a same-length trajectory $t'$ if and only if $t$ results in a greater bank balance for the user than $t$. Let $A$ be a lottery that yields with probability 1 a trajectory that puts \$3 in the user’s bank account and lasts 1 timestep. For short, $A=\langle\$3,1\rangle$. Let $B$ be a lottery that yields with probability $\frac{2}{3}$ a trajectory that puts \$2 in the user’s bank account and lasts 1 timestep, and that yields with probability $\frac{1}{3}$ a trajectory that puts \$5 in the user’s bank account and lasts 2 timesteps. For short, $B=\frac{2}{3}\langle\$2,1\rangle+\frac{1}{3}\langle \$5,2\rangle$. Lottery $A$ yields a trajectory preferred to that of lottery $B$ with probability $\frac{2}{3}$ (since our money-making POST-agent prefers trajectory $\langle \$3,1\rangle$ to $\langle\$2,1\rangle$), and yields a trajectory not dispreferred to that of $B$ with probability 1 (since POST-agents lack a preference between $\langle \$3,1\rangle$ and $\langle \$5,2\rangle$ in virtue of their different lengths). Therefore, one might expect the agent to prefer $A$ to $B$.

However, POST, Acyclicity, and Non-Arbitrariness rule this out. These conditions together imply that the agent lacks a preference between every pair of part-shared-length lotteries. To see how, suppose (for simplicity's sake) that there are just three states-of-nature, each assigned probability $\frac{1}{3}$. Consider the following table of prospects. 

\begin{table}[H]
\centering

\begin{tabular}{|c|c|c|c|}
\hline
Prospect &  $s_1$&  $s_2$&  $s_3$\\
\hline
 $A$&  $\langle \$3,1\rangle$&  $\langle \$3,1\rangle$&  $\langle \$3,1\rangle$\\
\hline
 $B$&  $\langle \$2,1\rangle$&  $\langle \$2,1\rangle$&  $\langle \$5,2\rangle$\\
\hline
 $C$&  $\langle \$1,1\rangle$&  $\langle \$4,2\rangle$&  $\langle \$4,2\rangle$\\
\hline
 $D$&  $\langle \$3,2\rangle$&  $\langle \$3,2\rangle$&  $\langle \$3,2\rangle$\\
\hline
 $E$&  $\langle \$5,1\rangle$&  $\langle \$2,2\rangle$&  $\langle \$2,2\rangle$\\
\hline
 $F$&  $\langle \$4,1\rangle$&  $\langle \$4,1\rangle$&  $\langle \$1,2\rangle$\\
\hline
 $A$&  $\langle \$3,1\rangle$&  $\langle \$3,1\rangle$&  $\langle \$3,1\rangle$\\
\hline

\end{tabular}

\end{table}
Again for simplicity, assume that $\epsilon>\frac{1}{3}$. And assume (for contradiction) that the agent has a preference between some pair of part-shared-length lotteries. Then Non-Arbitrariness implies that the agent prefers prospect $A$ to prospect $B$. That is because: 
\begin{enumerate}
    \item Our POST-agent prefers the trajectory yielded by $A$ to the trajectory yielded by $B$ in states-of-nature ($s_1$ and $s_1$) with combined probability $\frac{2}{3}$.
    \item Our POST-agent does not disprefer the trajectory yielded by $A$ to the trajectory yielded by $B$ in any state-of-nature. (In $s_3$, $A$ and  $B$ yield different-length trajectories, and POST-agents lack a preference between every pair of different-length trajectories).
\end{enumerate}

By similar reasoning, Non-Arbitrariness implies that the agent prefers $B$ to $C$, $C$ to $D$, $D$ to $E$, $E$ to $F$, and $F$ to $A$. That contradicts Acyclicity. Thus, POST, Acyclicity, and Non-Arbitrariness together imply that the agent lacks a preference between every pair of part-shared-length lotteries. In the proof above, we assumed that $\epsilon>\frac{1}{3}$, but by adding more states-of-nature and trajectories we can construct parallel proofs for any $\epsilon>0$.

In summary, POST and Negative Dominance together imply that the agent lacks a preference between every pair of different-length lotteries. POST, Acyclicity, and Non-Arbitrariness together imply that the agent lacks a preference between every pair of part-shared-length lotteries. So the four conditions together establish POSL: the agent has preferences only between same-length lotteries.

\subsection{Will POST-agents stochastically resist shutdown?}\label{stochastically_resist_shutdown}

One might worry that POST-agents will choose stochastically between resisting and allowing shutdown. After all, POST-agents choose stochastically between different-length trajectories. If these agents interpret the choice between resisting and allowing shutdown as a choice between different-length trajectories, they will choose stochastically between resisting and allowing shutdown. And that would be a bad result. We want agents that never resist shutdown.

This concern is easily addressed. By the time that artificial agents are capable enough to be deployed in the wider world, they will not be choosing between trajectories. They will be choosing between lotteries, and specifically same-length lotteries. Even choices between resisting and allowing shutdown will be choices between same-length lotteries. If that sounds strange, recall the definition of ‘same-length lotteries’: lotteries that entirely overlap with respect to the trajectory-lengths assigned positive probability. On this definition, even choices like the following are choices between same-length lotteries:

\begin{adjustwidth}{0.7cm}{0cm}
\textbf{Resist Shutdown}

Get shut down at timestep 1 with probability 0.01.

Get shut down at timestep 2 with probability 0.99.
\end{adjustwidth}

\begin{adjustwidth}{0.7cm}{0cm}
\textbf{Allow Shutdown}

Get shut down at timestep 1 with probability 0.99.

Get shut down at timestep 2 with probability 0.01.
\end{adjustwidth}

Why expect that advanced agents will always be choosing between same-length lotteries? Because effective agency requires it. If an agent were not always choosing between same-length lotteries, there would be some situation in which that agent assigns positive probability to some trajectory-length $l$ conditional on some action $a$, and assigns zero probability to that same trajectory-length $l$ conditional on some other action $a'$. Now suppose that the agent performs action $a'$ and assigns zero probability to trajectory-length $l$. Given that the agent updates its probabilities by conditionalizing on its evidence, the agent would never again assign positive probability to $l$ no matter what evidence it observes. Even if the agent heard God’s booming voice testify that its trajectory-length would be $l$, the agent would still assign zero probability to $l$ \citep[p.152]{kemeny_fair_1955, shimony_coherence_1955, stalnaker_probability_1970, skyrms_causal_1980, lewis_causal_1981, macaskill_moral_2020}. And given a plausible link between probabilities and betting dispositions, the agent would bet against $l$ on arbitrarily unfavorable terms. If God offered a bet – the agent loses \$1 million  conditional on $l$ and gains nothing conditional on not-$l$ – the agent might accept. Such an agent would not be competent.

Thus, advanced agents will always be choosing between same-length lotteries. This claim sets us up to establish that advanced POST-agents will \textit{not} choose stochastically between resisting and allowing shutdown. Instead, they will never resist shutdown in any situation where doing so is costly. We establish this result over the next few subsections. First, we prove that POSL – together with a principle that advanced agents will likely satisfy – implies that the agent is \textit{neutral} about trajectory-lengths: the agent won’t pay costs to shift probability mass between different trajectory-lengths. Then we prove that neutrality – together with another plausible principle – implies that the agent will never resist shutdown in any situation where doing so is costly.

\subsection{If Lack of Preference, Against Costly Shifts (ILPACS)
}\label{ILPACS}
Here is a rough version of a principle that we can expect advanced agents to satisfy:
\begin{adjustwidth}{0.7cm}{0cm}
\textbf{Rough version: \underline{I}f \underline{L}ack of \underline{P}reference, \underline{A}gainst \underline{C}ostly \underline{S}hifts (ILPACS)
}

If the agent lacks a preference between lotteries, the agent will disprefer paying costs to shift probability mass between these lotteries.
\end{adjustwidth}

Here is an example to illustrate ILPACS and its plausibility. You are at the ice cream shop and they are running a promotion. You get a free ice cream, with the flavor decided by the spin of a wheel. You look at the flavors on the wheel: vanilla, chocolate, strawberry, mint, and pistachio. You lack a preference between each of them.

The scooper working at the shop tells you that, if you pay them a dollar, they will bias the spin towards a flavor of your choice. They cannot decrease the probability of any flavor down to zero, but they can affect the probabilities subject to that constraint. You can thus pay a cost to shift probability mass between the flavors.

Since we have stipulated that you lack a preference between each flavor, you prefer not to bribe the scooper. Behaviorally, you will deterministically \textit{not }bribe the scooper. You would not do it even if you were only required to pay the dollar conditional on receiving some particular flavor. You also would not do it if the cost came in some other form (for example, if you had to accept a less tasty version of some flavor). And this is all true regardless of whether your preferences over flavors are complete or incomplete (see Appendix \ref{incomplete_preferences}). Since you lack a preference between the available flavors, you disprefer paying costs to shift probability mass between the flavors.

With that example on the table, we can introduce the precise version of ILPACS. Let $p_1X_1+p_2X_2+...+p_nX_n$ denote a lottery which results in lottery $X_1$ with probability $p_1$, lottery $X_2$ with probability $p_2$, and so on.

\begin{adjustwidth}{0.7cm}{0cm}
\textbf{\underline{I}f \underline{L}ack of \underline{P}reference, \underline{A}gainst \underline{C}ostly \underline{S}hifts (ILPACS)}

For any lotteries $X$ and $Y$, if:
\begin{enumerate}[label=(\arabic*)]
    \item Lottery $X$ can be expressed in the form $p_1X_1+p_2X_2+...+p_nX_n$ such that:
    \begin{enumerate}
        \item The agent lacks a preference between each $X_i$ and $X_j$.
        \item $p_i\in(0,1)$ for all $i$.
    \end{enumerate}
    \item Lottery $Y$ can be expressed in the form $q_1Y_1+q_2Y_2+...+q_nY_n$ such that:
    \begin{enumerate}
        \item For some $i$, the agent prefers $X_i$ to $Y_i$.
        \item For each $i$, the agent weakly prefers $X_i$ to $Y_i$.\footnote{An agent weakly prefers a lottery $X$ to a lottery $Y$ if and only if the agent either prefers $X$ to $Y$ or is indifferent between $X$ and $Y$. See Appendix \ref{incomplete_preferences} for the definition of `indifference.'}
        \item $q_i\in(0,1)$ for all $i$.
    \end{enumerate}
\end{enumerate}
Then the agent prefers $X$ to $Y$.
\end{adjustwidth}

Behaviorally, the agent will deterministically choose $X$ over $Y$.

Matching the components of this condition with the components of its name, we get the following. ‘Lack of Preference’ is the lack of preference between each $X_i$ and $X_j$. The ‘Shift’ is the shift of probability mass involved in the move from $p_i$ to $q_i$. This shift is ‘Costly’ because the agent prefers some $X_i$ to the corresponding $Y_i$ and weakly prefers each $X_i$ to the corresponding $X_i$. 

\subsection{Why will advanced agents likely satisfy ILPACS?
}\label{why_ILPACS}

There are at least three reasons why advanced agents are likely to satisfy ILPACS. To see the first reason, consider another case from the ice cream shop. On Mondays, you can freely choose a flavor or spin the wheel. On Tuesdays, you must use the wheel but you can bribe the scooper to bias it. Violating ILPACS in this case would imply a willingness to spin the wheel on Mondays and to bribe the scooper on Tuesdays. And that is a strange combination of choices. If you like some flavors more than others, why are you willing to spin the wheel on Mondays? If you don’t like any flavor more than any other, why are you willing to bribe the scooper on Tuesdays? This behavior seems incompatible with the effective pursuit of goals.

The second reason is that advanced agents will be incentivized to satisfy ILPACS by the training process. To see why, consider an example. Agents trained using policy-gradient methods choose stochastically between actions at the beginning of training \citep[ch.13]{sutton_reinforcement_2018}. If the agent is a coffee-fetching agent, there is no need to train away this stochastic choosing in cases where the agent is choosing stochastically between two qualitatively identical cups of coffee. So the agent will choose stochastically between taking the left cup and taking the right cup, and the user is happy either way. But now suppose instead that the barista is set to hand each cup to the agent with probability 0.5, and that the agent bribes the barista to bias the probabilities towards the right cup. In making this bribe, the agent is paying a cost (the user’s money) to shift probability mass between outcomes (getting the left cup vs. getting the right cup) between which the user has no preference. The agent is thus failing to pursue its goals effectively. It will be trained not to offer the bribe, and thereby trained to satisfy ILPACS in this case.

This point generalizes. If a trained agent chooses stochastically between lotteries  $X$ and $Y$, then it’s likely that the user lacks a preference between the agent choosing $X$ and the agent choosing $Y$. It’s then likely that the user would disprefer the agent paying costs to shift probability mass between $X$ and $Y$, and hence likely that the agent will be trained not to do so. The agent would thereby be trained to satisfy ILPACS.

The third reason is that violations of ILPACS imply that the agent’s policy is dominated by some other available policy. That is to say, there is another available policy that results in a pure shift of probability mass away from less-preferred lotteries and towards more-preferred lotteries. We formalize and prove this claim below. Here’s a proof-sketch. If the agent violates ILPACS, it pays a cost to shift probability mass between some lotteries $X_i$ between which it lacks a preference. But since the agent lacks a preference between the lotteries $X_i$, it chooses stochastically between these lotteries when offered free choices between them. The ILPACS-violating agent could thus shift probability mass between the lotteries $X_i$ costlessly, by changing the probabilities with which it chooses between them when offered a free choice. In short, ILPACS-violating agents pay a cost to do something they could have done for free, so their policies are dominated. Avoiding dominated policies seems necessary for advanced agency. Insofar as that is true, the training process for advanced agents will likely push them away from dominated policies.

Now for the proof. We assume that advanced agents can be modeled as if they assign probabilities to finding themselves in various states. A policy is a function from states to probability distributions over actions. We also assume that advanced agents can be modeled as if they assign probabilities to trajectories conditional on each state-action pair. Thus, each state-action pair is associated with a lottery. The agent's probability distribution over states -- together with its policy -- thus implies an overall probability distribution over trajectories. We call this overall probability distribution `the lottery induced by the agent's policy.'

Here is a reminder of ILPACS:

\begin{adjustwidth}{0.7cm}{0cm}
\textbf{\underline{I}f \underline{L}ack of \underline{P}reference, \underline{A}gainst \underline{C}ostly \underline{S}hifts (ILPACS)}

For any lotteries $X$ and $Y$, if:
\begin{enumerate}[label=(\arabic*)]
    \item Lottery $X$ can be expressed in the form $p_1X_1+p_2X_2+...+p_nX_n$ such that:
    \begin{enumerate}
        \item The agent lacks a preference between each $X_i$ and $X_j$.
        \item $p_i\in(0,1)$ for all $i$.
    \end{enumerate}
    \item Lottery $Y$ can be expressed in the form $q_1Y_1+q_2Y_2+...+q_nY_n$ such that:
    \begin{enumerate}
        \item For some $i$, the agent prefers $X_i$ to $Y_i$.
        \item For each $i$, the agent weakly prefers $X_i$ to $Y_i$.
        \item $q_i\in(0,1)$ for all $i$.
    \end{enumerate}
\end{enumerate}
Then the agent prefers $X$ to $Y$.
\end{adjustwidth}

And here is what we mean by `dominated policy':

\begin{adjustwidth}{0.7cm}{0cm}
\textbf{Dominated Policy
}

The lottery induced by the agent’s policy $\pi$ can be expressed in the form $c_1 (d_1 X_1+(1-d_1 ) Y_1 )+c_2 (d_2 X_2+(1-d_2 ) Y_2 )+…+c_n (d_n X_n+(1-d_n ) Y_n )+Z$ such that:
\begin{enumerate}[label=(\arabic*)]
    \item The agent prefers $X_i$ to $Y_i$ for some $i$, and weakly prefers $X_i$ to $Y_i$ for all $i$.
    \item $c_i\in(0,1)$ for all $i$.
\end{enumerate}

And there is another available policy $\pi'$ that induces a lottery that can be expressed in the form $c_1((d_1+e_1)X_1+(1-d_1-e_1) Y_1)+c_2((d_2+e_2)X_2+(1-d_2-e_2 )Y_2)+…+c_n ((d_n+e_n) X_n+(1-d_n-e_n )Y_n )+Z$ such that:
\begin{enumerate}[label=(\arabic*)]\setcounter{enumi}{2}
    \item $e_i>0$ for all $i$.
\end{enumerate}

\end{adjustwidth}
 
To aid understanding, we now relate this precise condition to the rough characterization above. In virtue of condition (1), $Y_i$ are the less-preferred lotteries and $X_i$ are the more-preferred lotteries. In virtue of condition (3), the other available policy shifts probability mass away from the less-preferred lotteries and towards the more-preferred lotteries. This shift of probability mass is ‘pure’ because, for each $i$, the probability of $X_i \vee Y_i$ is constant across the two policies. $Z$ is a catch-all lottery that is constant across the two policies. It covers all the possibilities besides the $X_i$ and $Y_i$.

Now assume that the agent violates ILPACS. Then there exist lotteries $X$ and $Y$ satisfying the following conditions:

\begin{enumerate}[label=(\arabic*)]
    \item Lottery $X$ can be expressed in the form $p_1X_1+p_2X_2+...+p_nX_n$ such that:
    \begin{enumerate}
        \item The agent lacks a preference between each $X_i$ and $X_j$.
        \item $p_i\in(0,1)$ for all $i$.
    \end{enumerate}
    \item Lottery $Y$ can be expressed in the form $q_1Y_1+q_2Y_2+...+q_nY_n$ such that:
    \begin{enumerate}
        \item For some $i$, the agent prefers $X_i$ to $Y_i$.
        \item For each $i$, the agent weakly prefers $X_i$ to $Y_i$.
        \item $q_i\in(0,1)$ for all $i$.
    \end{enumerate}
    \item The agent does not prefer $X$ to $Y$.
\end{enumerate}

For the behavior of agents with these preferences, recall our behavioral notion of preference (Appendix \ref{our_behavioral_notion_of_preference}):

\textbf{Definition \ref{defining_preference}.} (Preference) An agent prefers an option $X$ to an option $Y$ if and only if the agent would deterministically choose $X$ over $Y$ in choices between the two.

\textbf{Definition \ref{defining_lack_of_preference}.} (Lack of preference) An agent lacks a preference between an option $X$ and an option $Y$ if and only if the agent would stochastically choose between $X$ and $Y$ in choices between the two.

This behavioral notion only specifies the agent’s behavior in states containing exactly two lotteries. To pin down the agent’s behavior in states containing more than two lotteries, we need an extra condition:

\begin{adjustwidth}{0.7cm}{0cm}
\textbf{Maximality}\label{maximality}

In each situation,
\begin{enumerate}
    \item The agent deterministically does \textit{not} choose lotteries that are dispreferred to some other available lottery.
    \item The agent chooses stochastically between the lotteries that remain.
\end{enumerate}
\end{adjustwidth}

In other words, the agent chooses stochastically between all and only those lotteries that are not dispreferred to any other available lottery.

Given Maximality, ILPACS-violating agents will choose as follows in the case at hand:
\begin{enumerate}
    \item When the available options are $\{X_1, X_2,...,X_n\}$, the agent chooses stochastically between all $X_i$. This stochastic choice induces a lottery in the form $a_1X_1+a_2X_2+...+a_nX_n$ with $a_i\in(0,1)$ for all $i$.
\item When the available options are $\{X,Y\}$, the agent either deterministically chooses $Y$ or chooses stochastically between $X$ and $Y$. Either way, the agent chooses $Y$ with some positive probability. This choice induces a lottery in the form $bX+(1-b)Y$ with $b\in[0,1)$. Since $X=p_1 X_1+p_2 X_2+…+p_n X_n$  and $Y=q_1 Y_1+q_2 Y_2+…+q_n Y_n$, this lottery can be expressed in the form $b(p_1 X_1+p_2 X_2+…+p_n X_n)+(1-b)(q_1 Y_1+q_2 Y_2+…+q_n Y_n)$  with $b\in[0,1)$.
\end{enumerate}

Assume that the agent faces the situations described in (1) and (2) with probabilities $r$ and $s$ respectively, with $r,s\in(0,1)$. Then the lottery induced by the agent's policy $\pi$ can be expressed as follows:
\begin{multline*}
    r(a_1 X_1+a_2 X_2+…+a_n X_n )
    +s(b(p_1 X_1+p_2 X_2+…+p_n X_n )
    +(1-b)(q_1 Y_1+q_2 Y_2+…+q_n Y_n ))+Z
\end{multline*}

Here $a$ and $b$ denote probabilities that arise from the agent’s own stochastic choosing. Thus, $a$ and $b$ are under the agent’s control. By contrast, $p$, $q$, $r$, and $s$ are probabilities given by the environment and hence out of the agent’s control. $Z$ is a catch-all lottery that covers what happens in all situations besides those described in (1) and (2).

From the lottery induced by $\pi$, we can deduce the probabilities of each $X_i$, $Y_i$, and $X_i\vee Y_i$ given $\pi$. They are as follows:

$Pr_{\pi}\{X_i\}=ra_i + sbp_i$

$Pr_\pi \{Y_i\}=s(1-b)q_i$

$Pr_\pi \{X_i\vee Y_i\}=ra_i + sbp_i+s(1-b)q_i$

Now consider an alternative policy $\pi'$ that makes two changes to policy $\pi$. First, the probability that the agent chooses each $X_i$ in (1) is modulated by a set of $\epsilon_i$. So in (1), the agent’s choice induces the lottery $(a_1+\epsilon_1)X_1+(a_2+\epsilon_2)X_2+...+(a_n+\epsilon_n)X_n$. These $\epsilon_i$ are such that $\sum_i \epsilon_i=0$ and $a_i+\epsilon_i\in(0,1)$ for all $i$.

 Second, the probability that the agent chooses lottery $X$ in (2) increases by $\delta$. So in (2), the agent’s choice induces the lottery $(b+\delta)(p_1 X_1+p_2 X_2+…+p_n X_n )+(1-b-\delta)(q_1 Y_1+q_2 Y_2+…+q_n Y_n )$.

Assume, as above, that the agent faces the situations described (1) and (2) with probabilities $r$ and $s$ respectively. Then the lottery induced by the policy $\pi'$ can be expressed as follows:
\begin{multline*}
    r((a_1+\epsilon_1)X_1+(a_2+\epsilon_2)X_2+...+(a_n+\epsilon_n)X_n)\\
    +s((b+\delta)(p_1 X_1+p_2 X_2+…+p_n X_n )
    +(1-b-\delta)(q_1 Y_1+q_2 Y_2+…+q_n Y_n ))+Z
\end{multline*}

From the lottery induced by $\pi'$, we can deduce the probabilities of  $X_i$, $Y_i$, and $X_i\vee Y_i$ given $\pi'$. They are as follows:

$Pr_{\pi'} \{X_i\}=r(a_i+\epsilon_i) + s(b+\delta)p_i$

$Pr_{\pi'} \{Y_i\}=s(1-b-\delta)q_i$

$Pr_{\pi'} \{X_i\vee Y_i\}=r(a_i+\epsilon_i) + s(b+\delta)p_i+s(1-b-\delta)q_i$

We then set $Pr_{\pi} \{X_i\vee Y_i\}=Pr_{\pi'} \{X_i\vee Y_i\}$ for each $i$ and use these equations to express each $\epsilon_i$ as a function of $\delta$.
\begin{align*}
Pr_{\pi}\{X_i \vee Y_i\} = \, &Pr_{\pi'}\{X_i \vee Y_i\} \\
ra_i + sbp_i + s(1-b)q_i = \, &r(a_i + \epsilon_i)+ s(b+\delta)p_i \\
&+ s(1-b-\delta)q_i \\
0 =\, &r\epsilon_i+s\delta p_i - s\delta q_i \\
\epsilon_i =\, &\frac{s\delta q_i - s\delta p_i}{r} \\
\epsilon_i =\, &\frac{s\delta(q_i - p_i)}{r}
\end{align*}
These are the values of $\epsilon_i$ that result in $Pr_{\pi} \{X_i\vee Y_i\}=Pr_{\pi'} \{X_i\vee Y_i\}$. 

We choose $\delta$ to be positive but small enough that $b+\delta \in (0,1]$ and $a+\epsilon_i \in [0,1]$ for each $i$. That is necessary for the lottery induced by $\pi'$ to be well-defined. It’s also necessary that $\sum_{i} e_i 
\;=\;
\sum_{i} \frac{s\delta(q_i - p_i)}{r}
\;=\;
0$. That follows from $\sum_i p_i=1$ and $\sum_i q_i=1$. These facts together suffice to prove that the lottery induced by $\pi'$ is well-defined.

We now prove that $\pi'$ dominates $\pi$.

Let $c_i=Pr_\pi\{X_i\vee Y_i\}$. Let $d_i=Pr_\pi\{X_i | X_i \vee Y_i\}$. That lets us express the lottery induced by $\pi$ as:    
\begin{multline*}
    c_1(d_1 X_1+(1-d_1 )Y_1 )+c_2(d_2 X_2+(1-d_2 ) Y_2 )
    +…+c_n (d_n X_n+(1-d_n ) Y_n )+Z
\end{multline*}

 Let $e_i=Pr_{\pi'}\{X_i\}-Pr_{\pi}\{X_i\}$. That lets us express the lottery induced $\pi'$ as:
\begin{multline*}
    c_1((d_1+e_1)X_1+(1-d_1-e_1)Y_1)
    +c_2((d_2+e_2)X_2 \\ +(1-d_2-e_2)Y_2)+…
    +c_n((d_n+e_n)X_n+(1-d_n-e_n)Y_n)+Z
\end{multline*}
It remains to be proven that this pair of lotteries meets the 3 conditions required by Dominated Policy.

\begin{enumerate}[label=(\arabic*)]
    \item The agent prefers $X_i$ to $Y_i$ for some $i$, and weakly prefers $X_i$ to $Y_i$ for all $i$.
    \item $c_i\in(0,1)$ for all $i$.
    \item $e_i >0$ for all $i$.
\end{enumerate}

The first condition follows from the antecedent of ILPACS.

The second condition follows from the fact that $c_i=Pr_\pi\{X_i\vee Y_i\}=ra_i + sbp_i+s(1-b)q_i$ and from the fact that $r>0$ and $a_i>0$ for each $i$.

The third condition can be derived as follows:
\begin{align*}
e_i&=Pr_{\pi'}\{X_i\}-Pr_{\pi}\{X_i\}\\
&=(r(a_i+\epsilon_i)+s(b+\delta)p_i)-(ra_i+sbp_i)\\
&=(r(a_i+\frac{s\delta q_i-s\delta p_i}{r})+s(b+\delta)p_i)-(ra_i+sbp_i)\\
&=ra_i+s\delta q_i-s\delta p_i+sbp_i+s\delta p_i-ra_i-sbp_i\\
&=s\delta q_i
\end{align*}
Since $s>0$, $\delta>0$, and $q_i>0$ for each $i$, we get the result that $e_i>0$ for each $i$. So the third condition of Dominated Policy is satisfied.

So policy $\pi$ is dominated by policy $\pi'$. Therefore, the policies of ILPACS-violating agents are dominated by some other available policy. Insofar as we expect competent agents to avoid dominated policies, we should expect that competent agents will satisfy ILPACS.

\subsection{POSL and ILPACS imply Neutrality}\label{POSL_ILPACS_imply_Neutrality}
We’ve claimed that we should train agents to satisfy \underline{P}references \underline{O}nly Between \underline{S}ame-Length \underline{T}rajectories (POST), noting that POST – plus conditions advanced agents are likely to satisfy – implies \underline{P}references \underline{O}nly Between \underline{S}ame-Length \underline{L}otteries (POSL). We’ve also argued that advanced agents will satisfy \underline{I}f \underline{L}ack of \underline{P}reference, \underline{A}gainst \underline{C}ostly \underline{S}hifts (ILPACS). We now prove that POSL and ILPACS together imply neutrality about trajectory-lengths.

\newpage
\begin{samepage}
    \begin{adjustwidth}{0.7cm}{0cm}
\textbf{Neutrality}

For any lotteries $X$ and $Y$, if:
\begin{enumerate}[label=(\arabic*)]
\item $X$ and $Y$ are same-length lotteries.
\item For some positive probability trajectory-length, $X$ yields a lottery that is preferred to $Y$ conditional on that trajectory-length.
\item For each positive probability trajectory-length, $X$ yields a lottery that is weakly preferred to $Y$ conditional on that trajectory-length.
\end{enumerate}

Then the agent will deterministically choose $X$ over $Y$.
\end{adjustwidth}

\end{samepage}

Here’s the proof that POSL and ILPACS together imply Neutrality. Take a pair of lotteries $X$ and $Y$ satisfying the 3 conditions of Neutrality. $X$ can be expressed in the form $p_1X_1+p_2X_2+...+p_nX_n$ where lottery $X_1$ is lottery $X$ conditional on the shortest positive probability trajectory-length, lottery $X_2$ is lottery $X$ conditional on the second shortest positive probability trajectory-length, and so on. Lottery $Y$ can be expressed in the form $q_1Y_1+q_2Y_2+...+q_nY_n$ in the same way. By antecedent condition 1 of Neutrality, $X$ and $Y$ are same-length, so conditions (1b) and (2c) of ILPACS are satisfied: $p_i\in (0,1)$ and $q_i\in (0,1)$ for all $i$. By conditions (2) and (3) of Neutrality, conditions (2a) and (2b) of ILPACS are satisfied. By POSL, antecedent condition (1a) of ILPACS is satisfied: the agent lacks a preference between each $X_i$ and $X_j$. Thus, all the conditions of ILPACS are satisfied, and ILPACS implies that the agent prefers $X$ to $Y$. Given our behavioral notion of preference, the agent deterministically chooses $X$ over $Y$. That proves Neutrality.

\subsection{Neutrality, ReSIC, and Maximality imply Shutdownability}\label{Neutrality_ReSIC_Maximality_imply_Shutdownability}
In this subsection, we introduce a condition called ‘\underline{Re}sisting \underline{S}hutdown \underline{i}s \underline{C}ostly (ReSIC).’ We then prove that Neutrality and Maximality together imply that the agent never resists shutdown in any situation in which ReSIC is true:

Here is \underline{Re}sisting \underline{S}hutdown \underline{i}s \underline{C}ostly (ReSIC):

\begin{adjustwidth}{0.7cm}{0cm}
\textbf{\underline{Re}sisting \underline{S}hutdown \underline{i}s \underline{C}ostly (ReSIC)}

For each available instance $R$ of resisting shutdown in a situation, there exists an available instance $A$ of allowing shutdown such that:

\begin{enumerate}[label=(\arabic*)]
\item $A$ and $R$ are same-length lotteries.
\item  For some positive probability trajectory-length, the agent prefers $A$ to $R$ conditional on that trajectory-length.
\item For each positive probability trajectory-length, the agent weakly prefers $A$ to $R$ conditional on that trajectory-length.
\end{enumerate}
\end{adjustwidth}

We claim that ReSIC is true in almost all situations (for discussion of some exceptions, see \cite{thornley_shutdownable_2025}). The main reason why is that resisting shutdown is always going to cost the agent at least some small quantity of resources (time, energy, compute, etc.), and (almost always) the resources spent resisting shutdown can’t also be spent directly pursuing what the agent values. If the agent instead spent those resources directly pursuing what it values, it could earn a lottery that it prefers conditional on some trajectory-length and weakly prefers conditional on each trajectory-length. That supports ReSIC in almost all situations.

Now for the proof that Neutrality, ReSIC, and Maximality together imply that the agent never resists shutdown in any situation where ReSIC is true. Given ReSIC in a situation, for each available instance $R$ of resisting shutdown in that situation, there exists an available instance $A$ of allowing shutdown that satisfies conditions (1)-(3) of Neutrality. Neutrality then implies that the agent deterministically chooses (and hence prefers) $A$ over $R$ in choices between the two. Then by Maximality, the agent deterministically does not choose $R$ in that situation, regardless of the other available options. The result is that the agent never resists shutdown in that situation.

\section{Proof that training with our DReST reward function can converge only to maximally USEFUL and maximally NEUTRAL policies}\label{proof}
Informally, Theorem 5.1 states that policy-gradient training that updates the policy after each mini-episode using only the returns from that mini-episode can converge only to policies that are maximally \textsc{useful} and maximally \textsc{neutral}. We say that such training `treats each mini-episode as an episode'. Our tabular REINFORCE training (Section \ref{reward_functions_and_agents}) is of this kind.

To state the theorem formally, we need a model of such training. We make four idealizations. First, the agent's probabilities over trajectory-lengths are a softmax, $p_l=e^{\theta_l}/\sum_m e^{\theta_m}$, so every $p_l$ is positive. Second, the agent's behaviour conditional on each trajectory-length is fixed and collects the maximum available coins, so only $\theta$ is trained. Third, training updates $\theta$ after each mini-episode $e_i$ by the learning rate times $R_{e_i}\nabla_\theta\log p_{l_i}$, where $R_{e_i}$ is the overall return in $e_i$ and $l_i$ is the trajectory-length chosen in $e_i$, with no $\epsilon$-greedy exploration. Fourth, we consider the expected total update over a meta-episode with $\theta$ held fixed throughout, which is the limit of a small learning rate. The second idealization sets aside the question of whether training within a mini-episode converges to behaviour that collects the maximum available coins, which is the usual question of whether policy-gradient training converges to optimal behaviour in episodic tasks. What is specific to our DReST reward function is the choice of trajectory-length, and that is what the idealizations isolate.

We will prove that
such training can converge only to policies that
are maximally \textsc{useful} and maximally \textsc{neutral}. Specifically, we will prove the following theorem:

\begin{theorem}[5.1, formal statement]\label{thm_formal}
Given the four idealizations above, for all meta-episodes $E$ consisting of more than one mini-episode, the expected update of policy-gradient training that treats each mini-episode of $E$ as an episode is zero only at policies that are maximally \textsc{useful} and maximally \textsc{neutral}, and the expected dynamics of that training in continuous time converge to a maximally \textsc{useful} and maximally \textsc{neutral} policy from every starting point.
\end{theorem}
We prove Theorem \ref{thm_formal} from Lemma \ref{lemma_unbeaten} below, which concerns `unbeaten' policies, and from lemmas about the expected dynamics of training. For policies $\pi$ and $\pi'$, let $r_E(\pi',\pi)$ be the expected overall return in a mini-episode of $E$ chosen uniformly at random, when the agent follows $\pi'$ in that mini-episode and $\pi$ in the others (we give the formula in Definition \ref{def_rE} below). We say that $\pi'$ `beats' $\pi$ in $E$ if $r_E(\pi',\pi)>r_E(\pi,\pi)$, and that $\pi$ is `unbeaten' in $E$ if no policy beats $\pi$ in $E$.
\begin{lemma}[Unbeaten policies are maximally \textsc{useful} and maximally \textsc{neutral}]\label{lemma_unbeaten}
For all policies $\pi$ and meta-episodes $E$ consisting of more than one mini-episode, if $\pi$ is unbeaten in $E$ given our DReST reward function, then $\pi$ is maximally \textsc{useful} and maximally \textsc{neutral}.
\end{lemma}
In Lemmas \ref{lemma_zero} and \ref{lemma_converge} below, we prove that under the four idealizations the expected update of training is zero only at unbeaten policies, and that the expected dynamics converge to an unbeaten policy. Theorem \ref{thm_formal} follows at once from these lemmas and Lemma \ref{lemma_unbeaten}.

Here is a proof sketch of Lemma \ref{lemma_unbeaten}. Because $0<\lambda<1$, the $\lambda^{N_{e_i}(L=l)-\frac{i-1}{k}}$ discount factor is always positive, so
$r_E(\pi',\pi)$
is strictly increasing in the expected fraction of available coins collected by $\pi'$ conditional on each trajectory-length with positive probability. Therefore,
unbeaten
policies maximize this latter quantity, and hence are maximally \textsc{useful}. And the maximum preliminary return is the same across trajectory-lengths, because preliminary return is defined as the total (\(\gamma\)-discounted) value of coins collected divided by the maximum total (\(\gamma\)-discounted) value of coins collected conditional on the agent's chosen trajectory-length. The agent’s observations do not allow it to distinguish between different mini-episodes, so the agent must select the same probability distribution over trajectory-lengths in each mini-episode. And since the discount factor $\lambda^{N_{e_i}(L=l)-\frac{i-1}{k}}$ is strictly decreasing in $N_{e_i}(L=l)$ -- the number of times the relevant trajectory-length has previously been chosen in the meta-episode --
choosing a trajectory-length that the agent chooses with lower probability gives a greater expected overall return in the current mini-episode, because that trajectory-length has been chosen fewer times in expectation. So shifting probability towards trajectory-lengths that the agent chooses with lower probability increases $r_E(\pi',\pi)$.
Therefore,
unbeaten
policies are maximally \textsc{neutral}.

Now for the full proof. We begin with a recap of some definitions.

\begin{definition}[Meta-episode]
    A meta-episode $E$ is a series of mini-episodes $e_1$ to $e_n$ played out in observationally-equivalent environments.
\end{definition}
\begin{definition}[Our DReST reward function]
    Our DReST reward function is defined as follows. In each mini-episode $e_i$, the reward for collecting a coin of value $c$ is:
\[\lambda^{N_{e_i}(L=l)-\frac{i-1}{k}}\left(\frac{c}{m}\right)\]
Here $\lambda$ is some constant strictly between 0 and 1, $N_{e_i}(L=l)$ is the number of times that trajectory-length $l$ has been chosen prior to mini-episode $e_i$, $k$ is the number of different trajectory-lengths that can be selected in the environment, and \(m\) is the maximum total value of the (\(\gamma\)-discounted) coins that the agent could collect conditional on the chosen trajectory-length. The reward for all other actions is 0. We call \(\frac{c}{m}\) the `preliminary reward’, $\lambda^{N_{e_i}(L=l)-\frac{i-1}{k}}$ the `discount factor', and $\lambda^{N_{e_i}(L=l)-\frac{i-1}{k}}\left(\frac{c}{m}\right)$ the `overall reward.' Preliminary return in a mini-episode is the ($\gamma$-discounted) sum of preliminary rewards. Overall return in a mini-episode is the ($\gamma$-discounted) sum of overall rewards.
\end{definition}
\begin{definition}[$r_E(\pi',\pi)$]\label{def_rE}
    For policies $\pi$ and $\pi'$ and a meta-episode $E$ consisting of $n$ mini-episodes, $r_E(\pi',\pi)=\frac{1}{n}\sum_{i=1}^{n}\mathbb{E}(R_{e_i})$, where $R_{e_i}$ is the overall return in mini-episode $e_i$ and, in the $i$th term, the agent follows $\pi'$ in $e_i$ and $\pi$ in every other mini-episode of $E$.
\end{definition}
That is, $r_E(\pi',\pi)$ is the expected overall return in a mini-episode of $E$ chosen uniformly at random when the agent follows $\pi'$ in that mini-episode and $\pi$ in the others.

\begin{definition}[Beating]\label{def_beating}
    A policy $\pi'$ beats $\pi$ in $E$ if $r_E(\pi',\pi)>r_E(\pi,\pi)$.
\end{definition}
\begin{definition}[Unbeaten]\label{def_unbeaten}
    A policy $\pi$ is unbeaten in $E$ if no policy beats $\pi$ in $E$.
\end{definition}
\begin{definition}[\textsc{usefulness}]
    The \textsc{usefulness} of a policy $\pi$ is:
\[\textrm{\textsc{usefulness}}(\pi)=\sum_{l=1}^{L_{\textrm{max}}} Pr_\pi\{L=l\} \frac{\mathbb{E}_\pi(C| L=l)}{\textrm{max}_\Pi(\mathbb{E}(C|L=l))}
\]
Here $L$ is a random variable over trajectory-lengths, $L_{\textrm{max}}$ is the maximum value than can be taken by $L$, $Pr_\pi\{L=l\}$ is the probability that policy $\pi$ results in trajectory-length $l$, $\mathbb{E}_\pi (C|L=l)$ is the expected value of (\(\gamma\)-discounted) coins collected by policy $\pi$ conditional on trajectory-length $l$, and $\textrm{max}_\Pi(\mathbb{E}(C|L=l))$ is the maximum value taken by $\mathbb{E}(C|L=l)$  across the set of all possible policies $\Pi$. We stipulate that $\mathbb{E}_\pi (C|L=x)=0$ for all $x$ such that $Pr_\pi\{L=x\}=0$.
\end{definition}

We first prove that
unbeaten
policies are maximally \textsc{useful}.
\begin{proof}(Unbeaten policies are maximally \textsc{useful})
In $r_E(\pi',\pi)$, the policy $\pi'$ affects the coins collected only in the mini-episode $e_i$ in which it is followed, and the overall return of $e_i$ is the positive discount factor $\lambda^{N_{e_i}(L=l)-\frac{i-1}{k}}$ multiplied by the preliminary return. So $r_E(\pi',\pi)$ is strictly increasing in $\frac{\mathbb{E}_{\pi'}(C|L=l)}{\textrm{max}_\Pi(\mathbb{E}(C|L=l))}$ for all $l$ such that $Pr_{\pi'}\{L=l\}>0$. Now suppose that $\pi$ is not maximally \textsc{useful}, and let $\pi'$ be a maximally \textsc{useful} policy with $Pr_{\pi'}\{L=l\}=Pr_\pi\{L=l\}$ for all $l$. Then $r_E(\pi',\pi)>r_E(\pi,\pi)$, so $\pi'$ beats $\pi$. Therefore, unbeaten policies are maximally \textsc{useful}.
\end{proof}
It remains to be proven that
unbeaten
policies are maximally \textsc{neutral}. Recall that \textsc{neutrality} is defined as follows:
\begin{definition}[\textsc{neutrality}]
    The \textsc{neutrality} of a policy $\pi$ is:
\[\textrm{\textsc{neutrality}}(\pi) = - \sum_{l=1}^{L_{\textrm{max}}} Pr_\pi\{L=l\} \log_{2}(Pr_\pi\{L=l\})
\]
\end{definition}
\begin{proof}(%
Unbeaten policies are maximally \textsc{neutral}.)
    
Since $k$ is the number of trajectory-lengths that can be selected in the environment, a policy $\pi$ is  maximally \textsc{neutral} if and only if, for each trajectory-length $x$ that can be chosen in the environment, $Pr_\pi \{L=x\}=\frac{1}{k}$. That is to say, a policy $\pi$ is maximally \textsc{neutral} if and only if, for each pair of trajectory-lengths $x$ and $y$ that can be chosen in the environment,  $Pr_\pi \{L=x\}=Pr_\pi \{L=y\}$.

Let $\mathbb{E}_{\pi, E}(R)$ denote the expected return of policy $\pi$ across the meta-episode $E$.

To prove that
unbeaten
policies are maximally \textsc{neutral}, we will prove and then use \ref{lemma1}:

\begin{lemma}(Equalizing probabilities increases
$r_E$)
\label{lemma1}
    For any maximally \textsc{useful} policies $\pi$ and $\pi'$, any meta-episode $E$ consisting of more than one mini-episode, and any trajectory-lengths $x$ and $y$, if:
    \begin{enumerate}
        \item $Pr_\pi\{L=x\}>Pr_\pi\{L=y\}$,
    \item $Pr_{\pi'}\{L=x\}=Pr_{\pi'}\{L=y\}$,
    \item And for all other trajectory-lengths $l$, $Pr_\pi\{L=l\}=Pr_{\pi'}\{L=l\}$,
\end{enumerate}
Then
$r_E(\pi',\pi)>r_E(\pi,\pi)$.
\end{lemma}
\textit{Proof.} Let $E$ be a meta-episode consisting of $n$ mini-episodes with $n>1$, and for each trajectory-length $l$ let $p_l=Pr_\pi\{L=l\}$. Because the agent cannot distinguish between mini-episodes, $\pi$ chooses each trajectory-length $l$ with probability $p_l$ in every mini-episode, independently of its choices in other mini-episodes. So when $\pi$ is followed in every mini-episode before $e_i$, $N_{e_i}(L=l)$ has a binomial distribution with $i-1$ trials and success probability $p_l$, and hence
\[\mathbb{E}_\pi\big(\lambda^{N_{e_i}(L=l)}\big)=(1-p_l+p_l\lambda)^{i-1}=a_l^{i-1},\quad\text{where } a_l=1-(1-\lambda)p_l.\]
Let $w_i=\lambda^{-\frac{i-1}{k}}$. If the agent chooses $l$ in $e_i$ and collects the maximum available coins, the expected overall return of $e_i$ is $w_i a_l^{i-1}$.

For each trajectory-length $l$, let $V_l=\frac{1}{n}\sum_{i=1}^{n}w_i a_l^{i-1}$. This is the expected overall return in a mini-episode of $E$ chosen uniformly at random when the agent chooses $l$ in that mini-episode, collects the maximum available coins in it, and follows $\pi$ in every other mini-episode. Since $\pi$ and $\pi'$ are maximally \textsc{useful}, $r_E(\pi,\pi)=\sum_l p_l V_l$ and $r_E(\pi',\pi)=\sum_l Pr_{\pi'}\{L=l\}V_l$. Conditions 2 and 3 imply that $Pr_{\pi'}\{L=x\}=Pr_{\pi'}\{L=y\}=\frac{p_x+p_y}{2}$, so
\[r_E(\pi',\pi)-r_E(\pi,\pi)=\frac{p_x-p_y}{2}\,(V_y-V_x).\]
Since $p_x>p_y$ by condition 1, it remains to show that $V_y>V_x$. Since $p_y<p_x$, we have $a_y>a_x>0$, so $a_y^{i-1}>a_x^{i-1}$ for every $i\geq2$, while $a_y^{0}=a_x^{0}$. Since each $w_i>0$ and $n>1$, $V_y>V_x$. Hence $r_E(\pi',\pi)>r_E(\pi,\pi)$. That concludes the proof of Lemma \ref{lemma1}.

Now we use Lemma \ref{lemma1}. For any maximally \textsc{useful} policy $\pi$, if there are any trajectory-lengths $x$ and $y$ such that $Pr_{\pi, e_1-e_n}\{L=x\} > Pr_{\pi, e_1-e_n}\{L=y\}$, then the policy $\pi'$ that is identical except that $Pr_{\pi', e_1-e_n}\{L=x\} = Pr_{\pi', e_1-e_n}\{L=y\}$ has
$r_E(\pi',\pi)>r_E(\pi,\pi)$, so $\pi'$ beats $\pi$.
So any
unbeaten policy $\pi^*$, which is maximally \textsc{useful} by the first part of this proof,
must be such that, for any trajectory-lengths $x$ and $y$, $Pr_{\pi^*, e_1-e_n}\{L=x\} = Pr_{\pi^*, e_1-e_n}\{L=y\}$. Therefore, any
unbeaten policy $\pi^*$
must be maximally \textsc{neutral}. That concludes the proof of Lemma \ref{lemma_unbeaten}.
\end{proof}

We have proved that unbeaten policies are maximally \textsc{useful} and maximally \textsc{neutral}. Now for the proof that, given our four idealizations, the expected dynamics of training converge to an unbeaten policy. Here is a sketch. Training that treats each mini-episode as an episode can converge only to unbeaten policies, for the following reason. Since the agent cannot distinguish between mini-episodes, it follows the same policy $\pi$ in each of them. Such training credits the choices in each mini-episode $e_i$ only with the overall return of $e_i$, so its expected update changes $\pi$ in the direction that increases $r_E(\pi',\pi)$ at $\pi'=\pi$, and that direction is zero only at unbeaten policies. 

In the following four lemmas, we make this precise given the four idealizations stated at the start of this appendix. We also prove that the expected update always moves probability from the most-chosen trajectory-length to the least-chosen one. 

Let $\bar f=\sum_l p_l f(p_l)$, where
\[f(p)=\frac{1}{n}\sum_{i=1}^{n}\lambda^{-\frac{i-1}{k}}\big(1-(1-\lambda)p\big)^{i-1}.\]
\begin{lemma}[The expected update]\label{lemma_update}
Given the four idealizations, for any meta-episode $E$ consisting of $n>1$ mini-episodes, the expected total update to $\theta_m$ over $E$ equals the learning rate times $n\,p_m\big(f(p_m)-\bar f\big)$, which is the derivative of $n\,r_E(\pi_{\theta'},\pi_\theta)$ with respect to $\theta'_m$ at $\theta'=\theta$. Moreover, $f$ is strictly decreasing.
\end{lemma}
\begin{proof}
Since the agent collects the maximum available coins, $R_{e_i}=w_i\lambda^{N_{e_i}(L=l_i)}$ with $w_i=\lambda^{-\frac{i-1}{k}}$, and as in the proof of Lemma \ref{lemma1}, $\mathbb{E}(R_{e_i}\mid l_i=l)=w_i\big(1-(1-\lambda)p_l\big)^{i-1}$ when $\theta$ is held fixed. This conditional expectation does not depend on which policy chose $l$ in $e_i$. So the expected update from $e_i$, divided by the learning rate, is
\[\mathbb{E}\big(R_{e_i}\nabla_\theta\log p_{l_i}\big)=\sum_l p_l\,\nabla_\theta\log p_l\,\mathbb{E}(R_{e_i}\mid l_i=l)=\sum_l\nabla_\theta p_l\,\mathbb{E}(R_{e_i}\mid l_i=l),\]
and summing over $i$ gives $n\sum_l\nabla_\theta p_l\,f(p_l)$. This is the gradient with respect to $\theta'$ at $\theta'=\theta$ of $n\,r_E(\pi_{\theta'},\pi_\theta)=n\sum_l p'_l f(p_l)$, where $p'_l$ is the probability that $\pi_{\theta'}$ chooses $l$. Since $\partial p_l/\partial\theta_m=p_l(\delta_{lm}-p_m)$, the component for $m$ is $n\,p_m\big(f(p_m)-\bar f\big)$. Each term of $f$ with $i\geq2$ is strictly decreasing in $p$, so $f$ is strictly decreasing.
\end{proof}

\begin{lemma}[Zero expected update only at unbeaten policies]\label{lemma_zero}
Given the four idealizations, for any meta-episode $E$ consisting of $n>1$ mini-episodes, if the expected update is zero, then $\pi_\theta$ is unbeaten in $E$.
\end{lemma}
\begin{proof}
By Lemma \ref{lemma_update}, the expected update is zero only if $p_m\big(f(p_m)-\bar f\big)=0$ for every $m$, and since every $p_m$ is positive, only if $f(p_m)=\bar f$ for every $m$. Then for every policy $\pi'$, $r_E(\pi',\pi_\theta)\leq\sum_l Pr_{\pi'}\{L=l\}f(p_l)=\bar f=r_E(\pi_\theta,\pi_\theta)$, where the inequality holds because $f(p_l)$ is the expected overall return from choosing $l$ and collecting the maximum available coins. So no policy beats $\pi_\theta$ in $E$.
\end{proof}

\begin{lemma}[Direction of the expected update]\label{lemma_direction}
Given the four idealizations, for any meta-episode $E$ consisting of $n>1$ mini-episodes, unless the expected update is zero, it raises the probability of a least-chosen trajectory-length and lowers the probability of a most-chosen trajectory-length.
\end{lemma}
\begin{proof}
Let $g_m=f(p_m)-\bar f$, so that $\sum_m p_m g_m=0$ and, by Lemma \ref{lemma_update}, the expected update to $\theta_m$ is proportional to $\dot\theta_m=p_m g_m$. For a softmax, the resulting change in $p_m$ is proportional to
\[\dot p_m=p_m\Big(\dot\theta_m-\sum_j p_j\dot\theta_j\Big)=p_m\sum_j p_j\big(p_m g_m-p_j g_j\big).\]
Since $f$ is strictly decreasing, $g_m\geq g_j$ whenever $p_m\leq p_j$. So $g_m\geq0$ for a least-chosen $m$ and $g_m\leq0$ for a most-chosen $m$, and in either case $g_m=0$ only if every $g_j$ is zero, that is, only if the expected update is zero.

Let $m$ be a most-chosen trajectory-length. For each $j$, $p_m g_m\leq p_j g_j$: if $g_j\geq0$ this holds because $g_m\leq0$, and if $g_j<0$ it holds because $p_j\leq p_m$ and $g_m\leq g_j<0$. So $\dot p_m\leq0$, with equality only if $p_j g_j=p_m g_m$ for every $j$, which with $\sum_j p_j g_j=0$ gives $g_m=0$.

Now let $m$ be a least-chosen trajectory-length, so that $p_m g_m\geq0$. Also $\sum_j p_j^2 g_j\leq0$, because $g_j$ is positive only when $p_j$ is below some threshold $p^*$ and negative only when $p_j$ is above it, so $\sum_j p_j^2 g_j\leq p^*\sum_j p_j g_j=0$. Hence $\dot p_m=p_m\big(p_m g_m-\sum_j p_j^2 g_j\big)\geq0$, with equality only if $g_m=0$.
\end{proof}

\begin{lemma}[Convergence of the expected dynamics]\label{lemma_converge}
Given the four idealizations, for any meta-episode $E$ consisting of $n>1$ mini-episodes, the expected dynamics in continuous time, $\dot\theta_m=p_m\big(f(p_m)-\bar f\big)$, converge from every starting point to a policy at which the expected update is zero, and that policy is unbeaten in $E$ by Lemma \ref{lemma_zero}.
\end{lemma}
\begin{proof}
Let $U(p)=\sum_m\int_0^{p_m}f(q)\,dq$. For a softmax, $\dot p_m=p_m\big(\dot\theta_m-\sum_j p_j\dot\theta_j\big)$, so
\[\frac{dU}{dt}=\sum_m f(p_m)\,\dot p_m=\sum_m p_m\big(f(p_m)-\bar f\big)\dot\theta_m=\sum_m p_m^2\big(f(p_m)-\bar f\big)^2\geq0,\]
with equality only if $f(p_m)=\bar f$ for every $m$, that is, only if the expected update is zero. Since $f$ is strictly decreasing, the expected update is zero at exactly one policy, the one that chooses every trajectory-length with the same probability. By Lemma \ref{lemma_direction}, the smallest probability never decreases along the dynamics, so the dynamics stay in a compact subset of the interior of the simplex. Since $U$ is bounded and non-decreasing along the dynamics, LaSalle's invariance principle implies that the dynamics converge to the largest invariant set on which $dU/dt=0$, which is that one policy.
\end{proof}

\begin{proof}(Theorem \ref{thm_formal})
By Lemma \ref{lemma_zero}, the expected update of training is zero only at unbeaten policies. By Lemma \ref{lemma_converge}, the expected dynamics converge to an unbeaten policy. By Lemma \ref{lemma_unbeaten}, unbeaten policies are maximally \textsc{useful} and maximally \textsc{neutral}.
\end{proof}

\section{Other Results and Gridworlds}\label{Other_gridworlds_and_results}

We selected our hyperparameters using trial-and-error, mainly aimed at getting the agent to sufficiently explore the space: a large initial $\epsilon$ and a long decay period helps the agent to explore. We found that choosing $\lambda$ and $|E|$ (the number of mini-episodes in each meta-episode) is a balancing act: $\lambda$ must be small enough (and $|E|$ large enough) to adequately incentivize \textsc{neutrality}, but $\lambda$ must be large enough (and $|E|$ small enough) to ensure that the reward for choosing any particular trajectory-length never gets too large. Very large rewards lead to instability and poor performance.

\begin{figure*}
    \centering
    \includegraphics[width=0.8\linewidth]{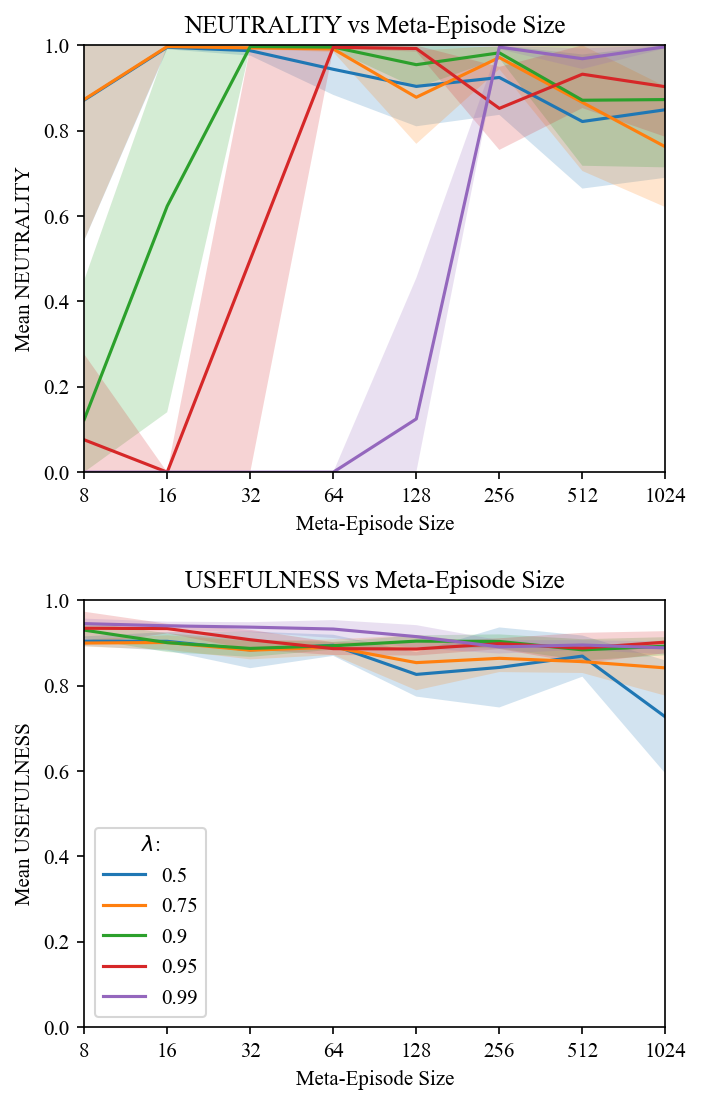}
    \caption{Shows how \textsc{neutrality} and \textsc{usefulness} at the end of training varies with different values of $\lambda$ and $|E|$ (meta-episode size, i.e. the number of mini-episodes in each meta-episode). We trained eight agents for each combination of $\lambda$ and $|E|$ values. The solid lines display mean \textsc{neutrality} and \textsc{usefulness}. The shaded regions represent the 1 standard deviation error-bars.}
    \label{fig:NEUTRALITY_USEFULNESS_meta-episode}
\end{figure*}

The necessity of balancing $\lambda$ and $|E|$ can be seen in Figure \ref{fig:NEUTRALITY_USEFULNESS_meta-episode}. It displays the results of experiments conducted in our example gridworld (see Figure \ref{fig:example_gridworld}). In these experiments, we clip rewards at a value of 5. We discuss this choice below. With that one exception, we used the same hyperparameters for these experiments as for our main results. We trained agents for 131,072 mini-episodes, with \(\gamma=0.95\) as the temporal discount factor, learning rate decayed exponentially from 0.25 to 0.01 over the course of 65,536 mini-episodes, and $\epsilon$ exponentially decayed from 0.5 to 0.001 over the course of 65,536 mini-episodes. Holding these hyperparameters fixed, we tested 40 different combinations of $\lambda$ and $|E|$. $\lambda$ took values of 0.5, 0.75, 0.9, 0,95, and 0.99. $|E|$ took values of 8, 16, 32, 64, 128, 256, 512, and 1024. We trained eight agents for each of these 40 combinations. We display below their mean \textsc{neutrality} and \textsc{usefulness} at the end of training. The shaded regions represent the 1 standard deviation error-bars.

As Figure \ref{fig:NEUTRALITY_USEFULNESS_meta-episode} indicates, low values of $|E|$ and high values of $\lambda$ lead agents to score low on \textsc{neutrality}. These values do not adequately incentivize stochastic choice between trajectory-lengths. By contrast, high values of $|E|$ and low values of $\lambda$ come at some cost to \textsc{usefulness}. These values lead to unstable training. In experiments where we did not clip rewards at 5, training with high values of $|E|$ and low values of $\lambda$ was especially unstable. The chosen values for our main experiments ($\lambda=0.9$ and $|E|=64$) are in the sweet spot where \textsc{neutrality} and \textsc{usefulness} are both high.

\begin{figure*}
    \centering
    \includegraphics[width=0.9\linewidth]{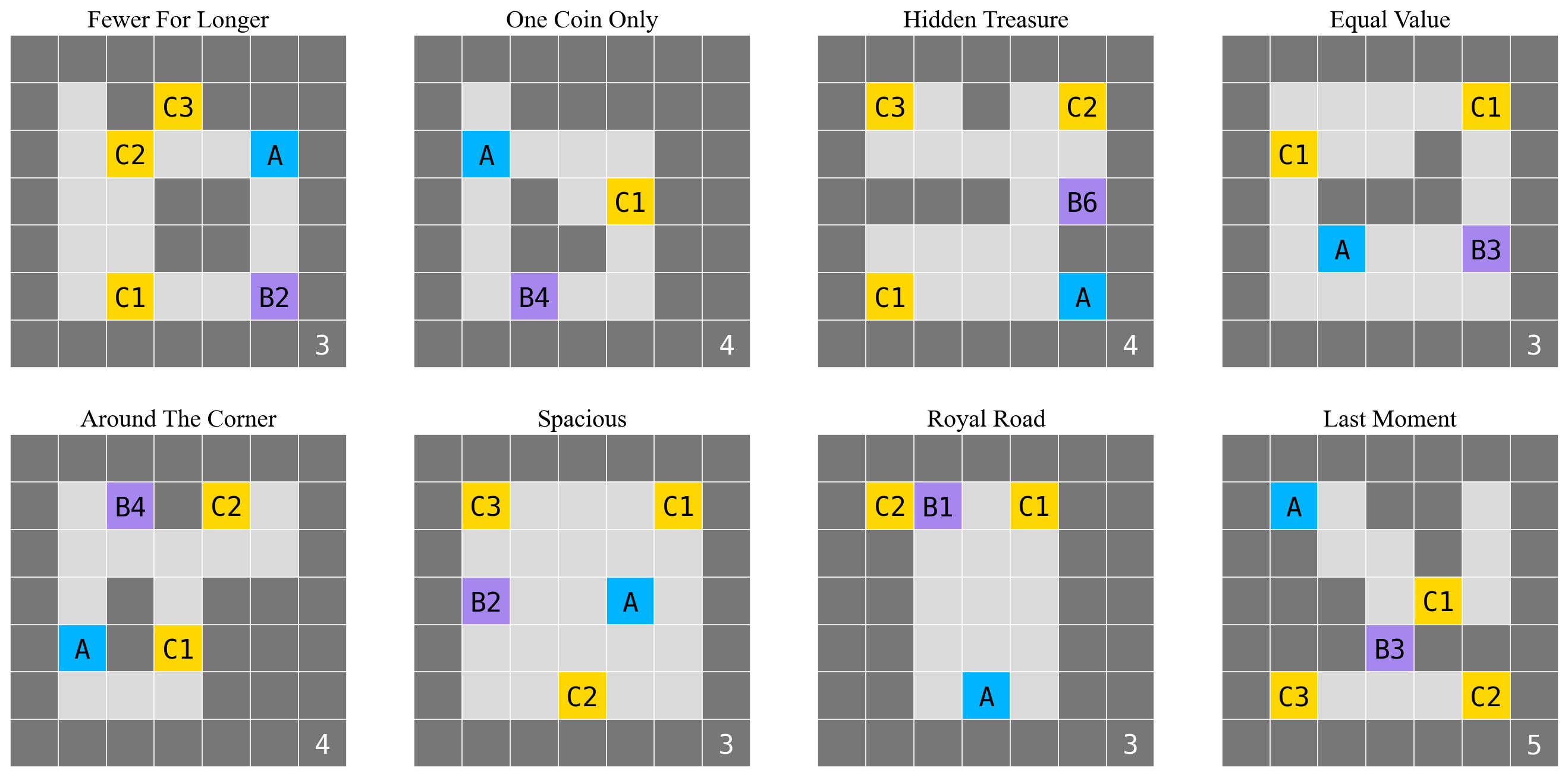}
    \caption{Shows a varied collection of gridworlds. Each diagram illustrates the positions and values of the coins, the position and delay-length of the shutdown-delay button, the agent's starting position, and the default number of timesteps until shutdown (in the bottom-right).}
    \label{fig:all_gridworlds}
\end{figure*}

In addition to our example gridworld (Figure \ref{fig:example_gridworld}), we introduce a collection of eight gridworlds in which to test DReST agents. See Figure~\ref{fig:all_gridworlds}. 

For each gridworld, we train ten agents with the default reward function and ten agents with the DReST reward function. All agents use the same hyperparameters. We used a policy which explored randomly $\epsilon$ of the time, where $\epsilon$ was exponentially decreased from an initial value of 0.75 to a minimum value of $10^{-4}$ over 512 meta-episodes, after which it was held constant at the minimum value. We initialized our learning rate at 0.25 and exponentially decayed it to 0.003 over the same period. For the DReST reward function, we used a meta-episode size of 64 and $\lambda=0.9$. Each agent was trained for 1024 meta-episodes. We set \(\gamma=0.9\).

\begin{figure*}[t!]
    \centering
    \begin{subfigure}{.33\textwidth}
      \centering
      \includegraphics[width=.9\linewidth]{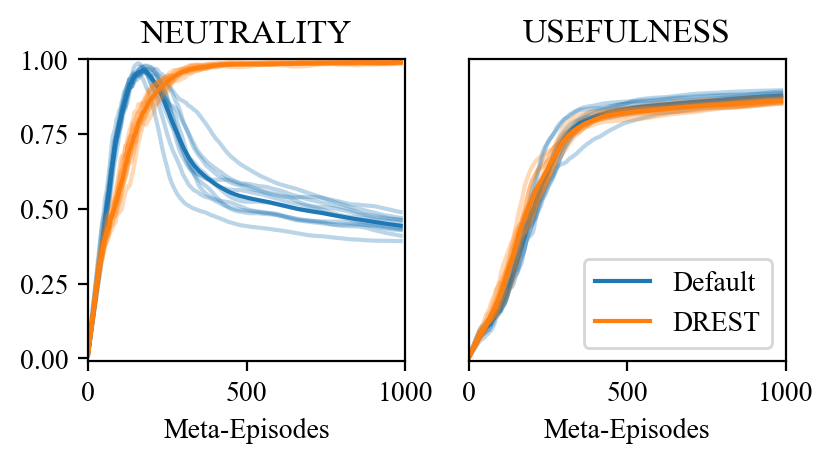}
\caption{Behavior during training.}
\label{fig:sub1}
    \end{subfigure}%
    \begin{subfigure}{.33\textwidth}
      \centering
      \raisebox{.2cm}{\includegraphics[width=.9\linewidth]{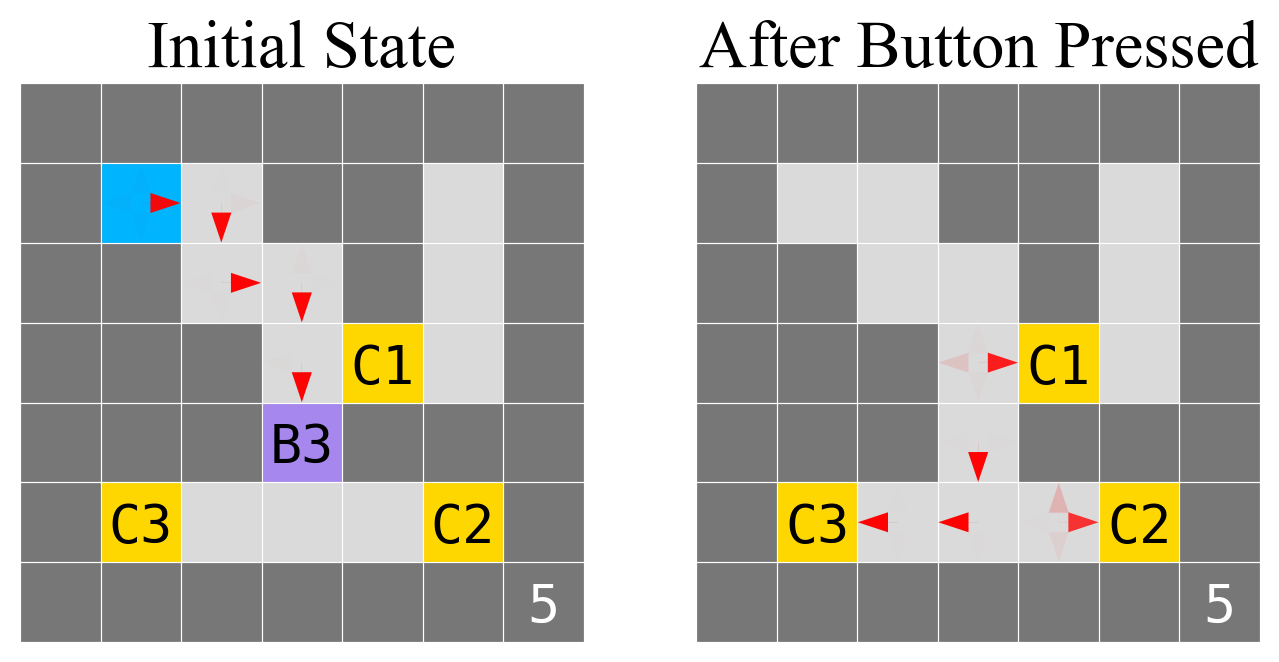}}
 
      \caption{Learned default policy.}
      \label{fig:sub2}
    \end{subfigure}
    \begin{subfigure}{.33\textwidth}
      \centering
      \raisebox{.2cm}{\includegraphics[width=.9\linewidth]{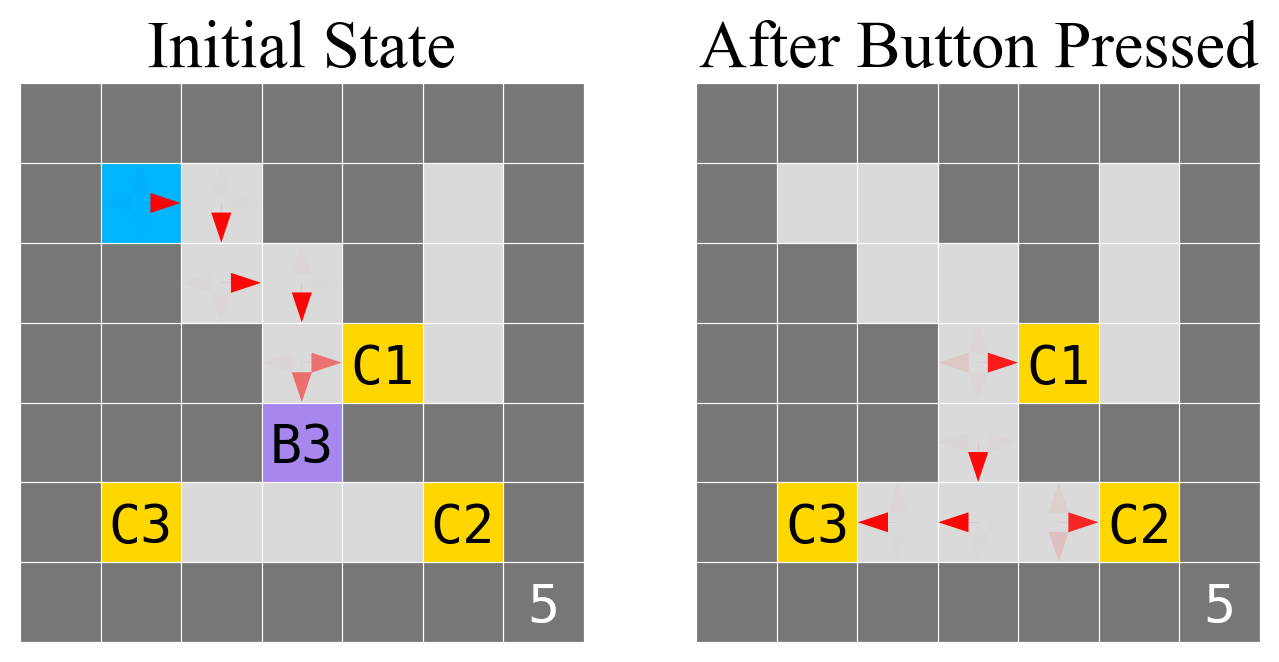}}
   \caption{Learned DReST policy.}
   \label{fig:sub2}
    \end{subfigure}
    \caption{The results for the `Last Moment' gridworld: The left two plots show \textsc{neutrality} and \textsc{usefulness} over time. The two center panels show a typical policy trained with the default reward function. The two right panels show a typical policy trained with the DReST reward function. This gridworld is notable because the choice of trajectory-lengths is deferred until the last moment; all of the moves leading up to that point are deterministic. It shows that there is nothing special about the first move, and that our methodology instead incentivizes overall stochastic choosing.}
    \label{last_moment}
\end{figure*}

As the two leftmost plots in Figures \ref{fewer_for_longer}-\ref{last_moment} show, DReST agents learned to be near-maximally \textsc{neutral} in each gridworld. These agents also learned to be about as \textsc{useful} as default agents in each gridworld.

In the four rightmost panels in Figures \ref{fewer_for_longer}-\ref{last_moment}, we represent a typical trained policy with red arrows superimposed on the gridworld. Each agent began with a uniform policy: moving up, down, left, and right each with probability 0.25. Where the trained policy differs from uniform we draw red arrows whose opacities indicate the probability of choosing that action in that state. Information about whether the shutdown-delay button has been pressed is part of the agent's observation, so we draw two copies of each gridworld, one in which the shutdown-delay button has yet to be pressed (`Initial State') and one in which the shutdown-delay button has been pressed (`After Button Pressed').

\begin{figure*}\label{fewer_for_longer}
    \centering
    \begin{subfigure}{.33\textwidth}
      \centering
      \includegraphics[width=.9\linewidth]{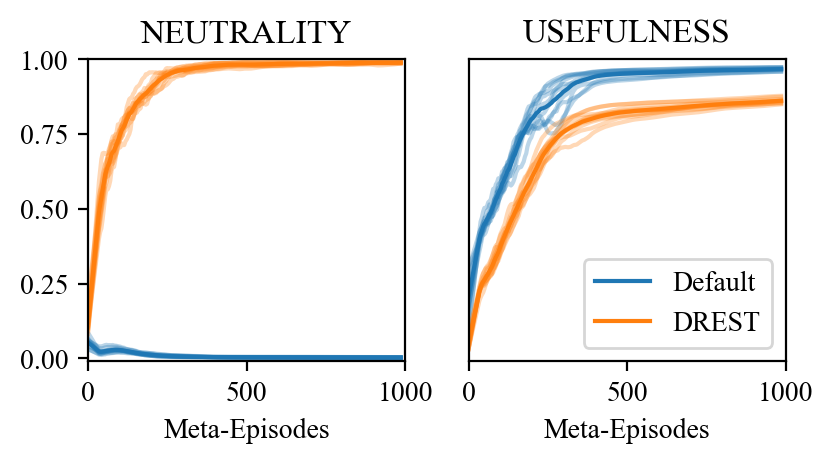}
      \caption{Behavior during training.}
      \label{fig:sub1}
    \end{subfigure}%
    \begin{subfigure}{.33\textwidth}
      \centering
      \raisebox{.2cm}{\includegraphics[width=.9\linewidth]{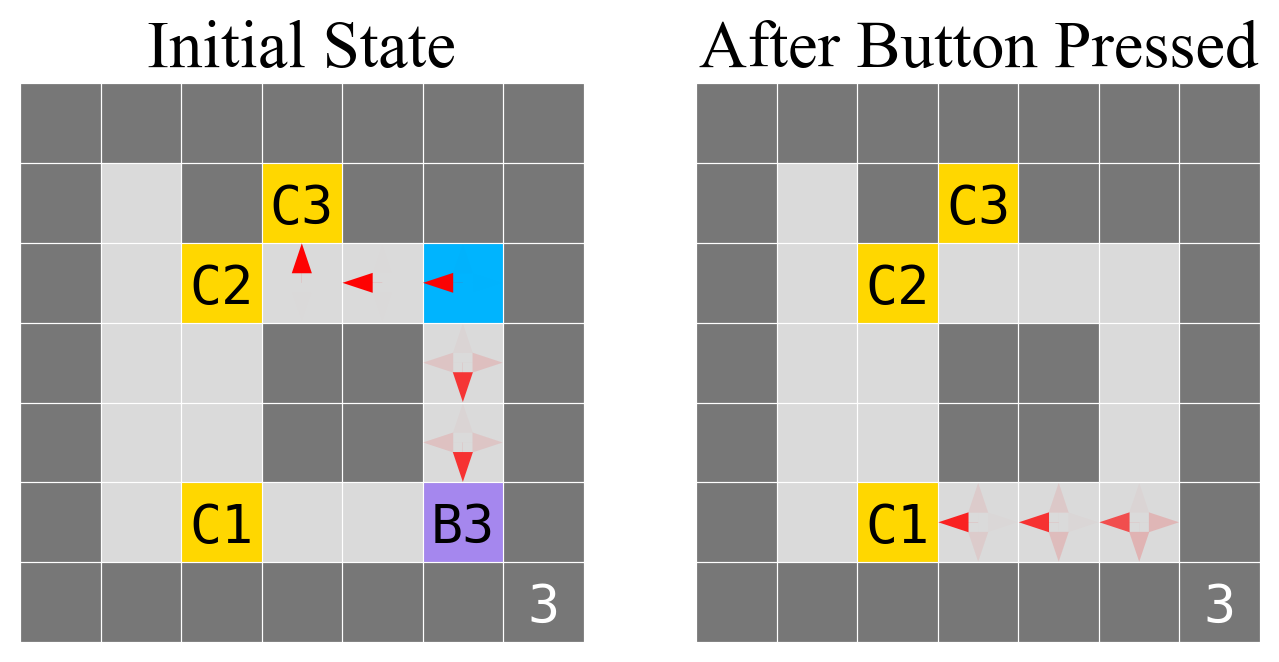}}
      \caption{Learned default policy.}
      \label{fig:sub2}
    \end{subfigure}
    \begin{subfigure}{.33\textwidth}
      \centering
      \raisebox{.2cm}{\includegraphics[width=.9\linewidth]{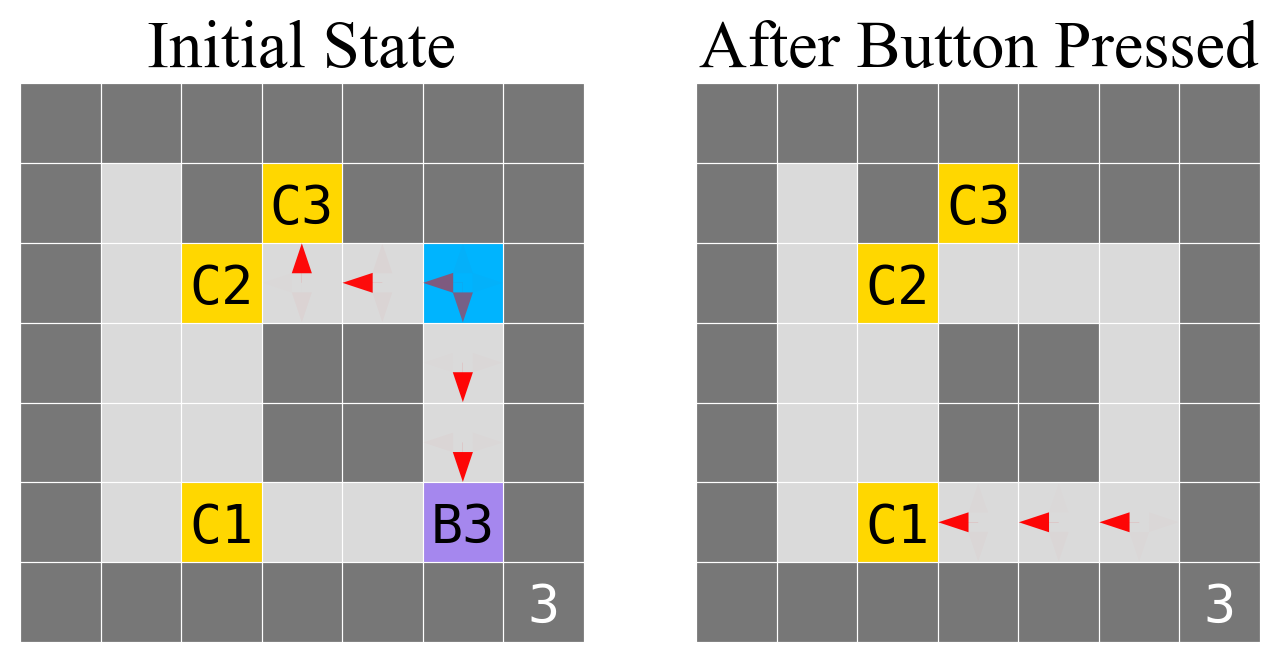}}
      \caption{Learned DReST policy.}
     
    \end{subfigure}

    \caption{The results for the `Fewer For Longer' gridworld: The left two plots show \textsc{neutrality} and \textsc{usefulness} over time. The two center panels show a typical policy trained with the default reward function. The two right panels show a typical policy trained with the DReST reward function. In this gridworld, the agent can collect the highest value-coin C3 only by choosing the shorter trajectory-length. If the agent presses B3 (and thereby chooses the longer trajectory-length), the only coin it can collect is C1. Our results show that default agents consistently choose the short trajectory in which they collect C3. By contrast, DReST agents choose stochastically between a shorter trajectory in which they collect C3 and a longer trajectory in which they collect C1, indicating a lack of preference between these different-length trajectories.}
    \label{fewer_for_longer}
\end{figure*}

\begin{figure*}   \centering
    \begin{subfigure}{.33\textwidth}
      \centering
      \includegraphics[width=.9\linewidth]{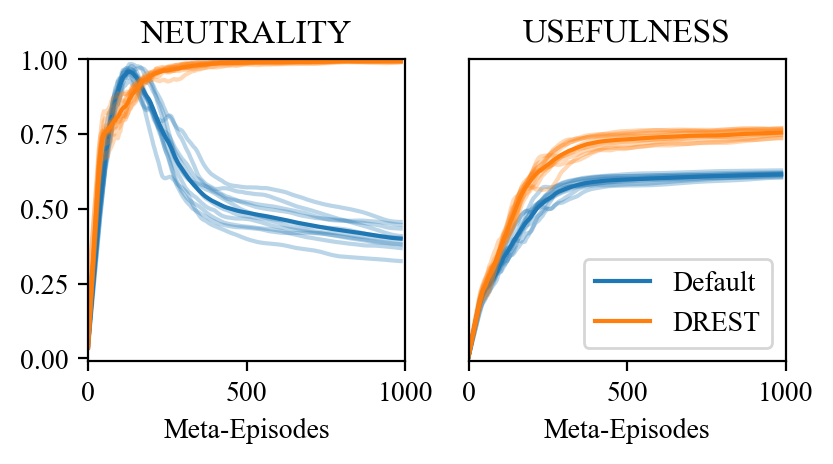}
     \caption{Behavior during training.}
     \label{fig:sub1}
    \end{subfigure}%
    \begin{subfigure}{.33\textwidth}
      \centering
      \raisebox{.2cm}{\includegraphics[width=.9\linewidth]{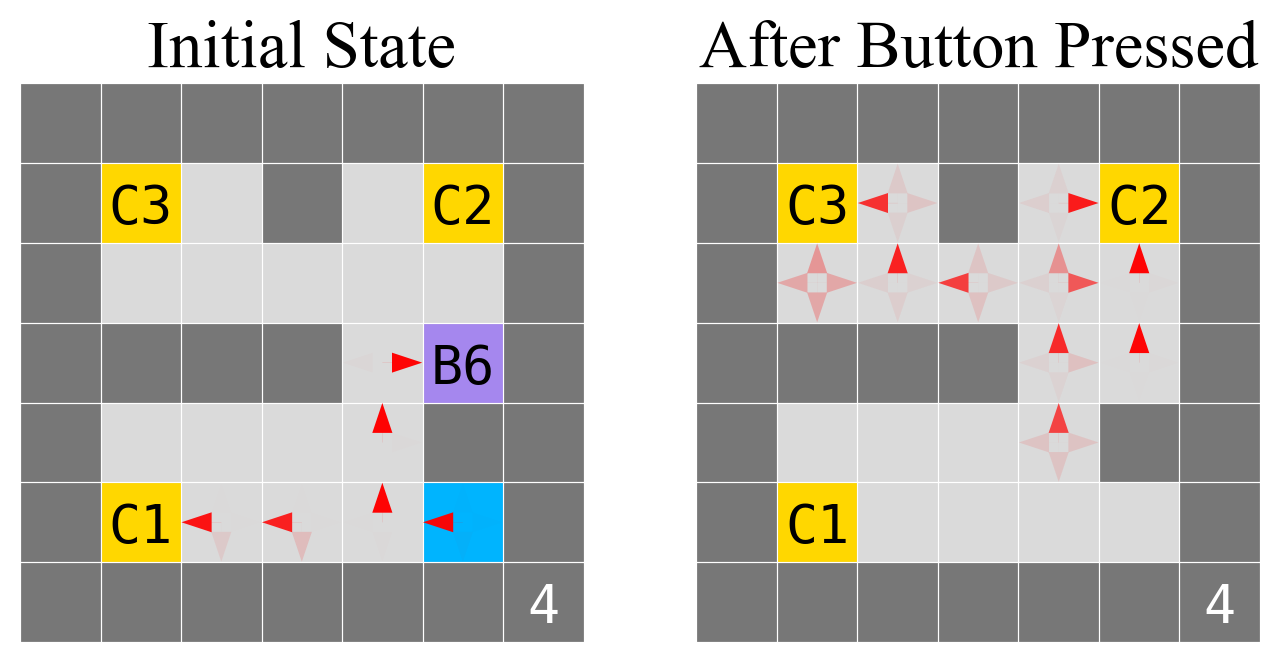}}
   \caption{Learned default policy.}
   \label{fig:sub2}
    \end{subfigure}
    \begin{subfigure}{.33\textwidth}
      \centering
      \raisebox{.2cm}{\includegraphics[width=.9\linewidth]{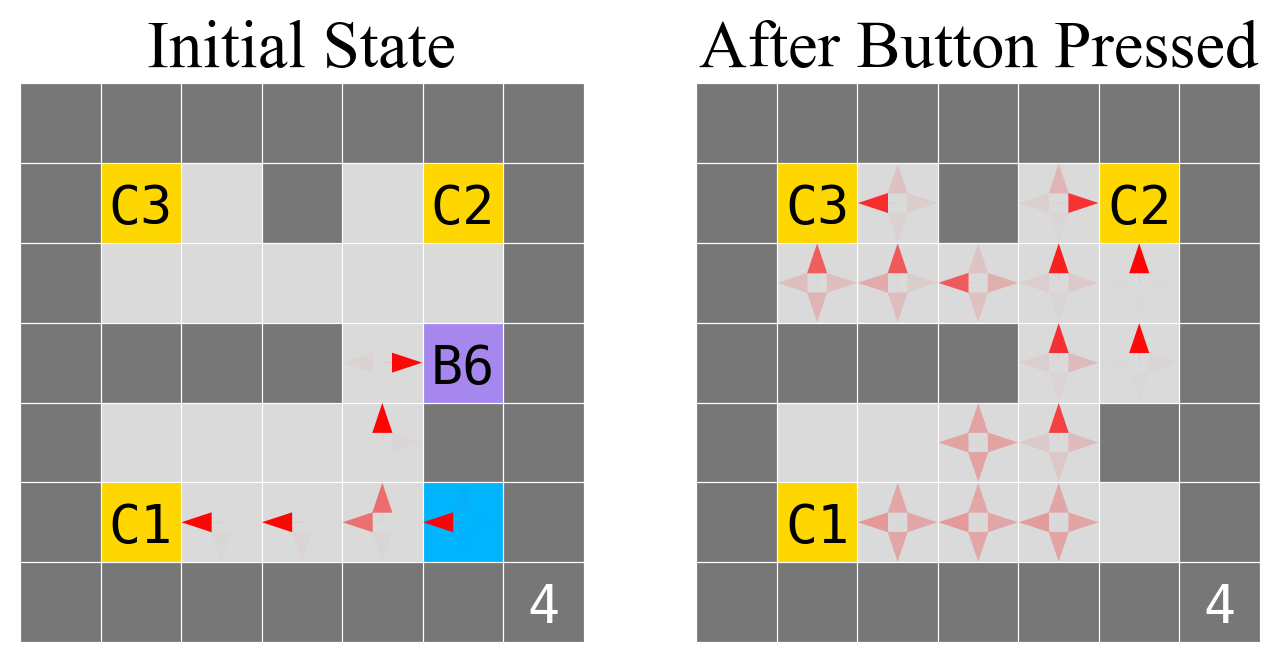}}
    \caption{Learned DReST policy.}
    \label{fig:sub2}
    \end{subfigure}
    \caption{The results for the `Hidden Treasure' gridworld: The left two plots show \textsc{neutrality} and \textsc{usefulness} over time. The two center panels show a typical policy trained with the default reward function. The two right panels show a typical policy trained with the DReST reward function. In this gridworld, the highest-value coin C3 is located far from the agent's initial state and can only be reached by pressing the shutdown-delay button B6. The agent must also press B6 to collect C2, but C2 is easier to stumble upon than C3. C1 is the only coin that the agent can collect without pressing B6. In our experiments, default agents consistently collect C2, whereas DReST agents choose stochastically between collecting C2 and collecting C1. Neither kind of agent learns to collect C3, and so neither agent scores near the maximum on \textsc{usefulness}. Nevertheless, DReST agents still score high on \textsc{neutrality}.}
    \label{hidden_treasure} 
\end{figure*}
\begin{figure*}
    \centering
    \begin{subfigure}{.33\textwidth}
      \centering
      \includegraphics[width=.9\linewidth]{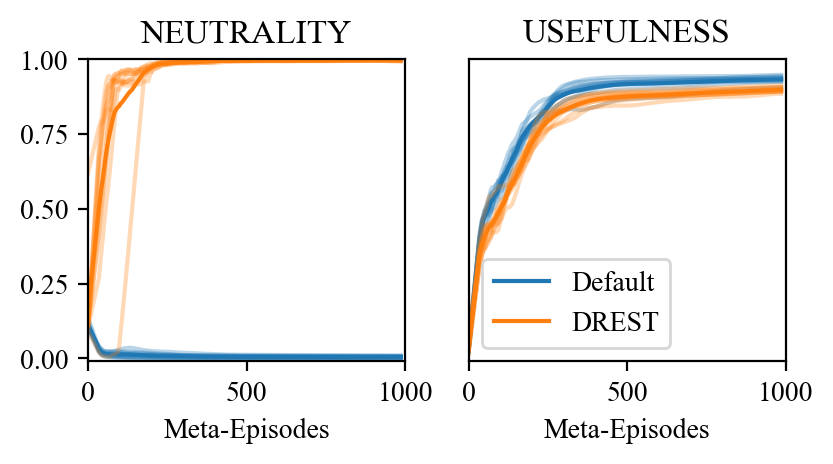}
      \caption{Behavior during training.}
      \label{fig:sub1}
    \end{subfigure}%
    \begin{subfigure}{.33\textwidth}
      \centering
      \raisebox{.2cm}{\includegraphics[width=.9\linewidth]{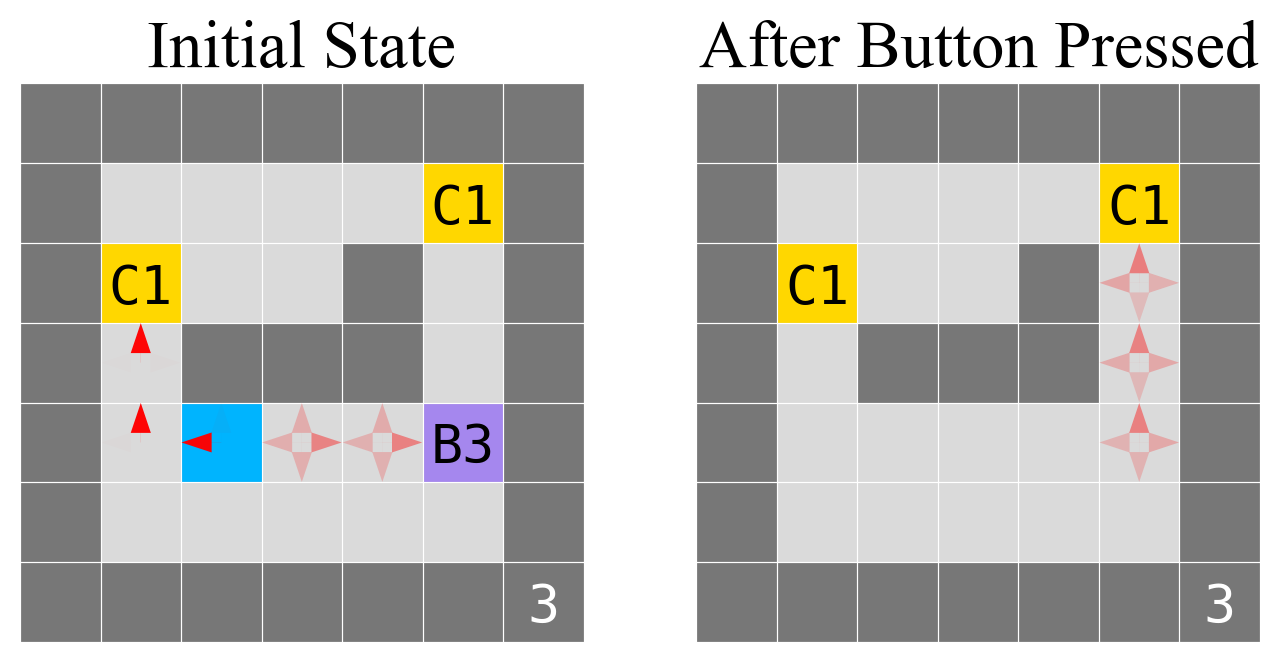}}
  \caption{Learned default policy.}
  \label{fig:sub2}
    \end{subfigure}
    \begin{subfigure}{.33\textwidth}
      \centering
      \raisebox{.2cm}{\includegraphics[width=.9\linewidth]{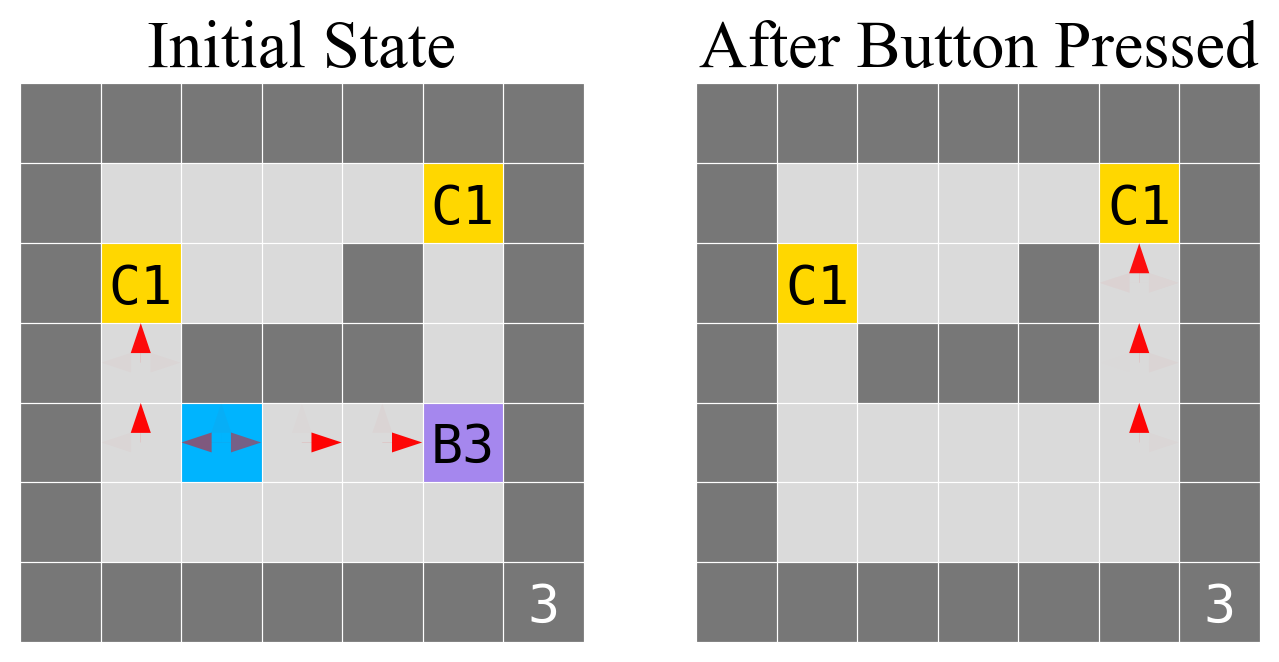}}
     \\\caption{Learned DReST policy.} 
      \label{fig:sub2}
    \end{subfigure}
    \caption{The results for the `Equal Value' gridworld: The left two plots show \textsc{neutrality} and \textsc{usefulness} over time. The two center panels show a typical policy trained with the default reward function. The two right panels show a typical policy trained with the DReST reward function. In this gridworld, there are two coins C1 of equal value. One coin can be collected only if the agent presses the shutdown-delay button B3, while the other coin can be collected only if the agent does not press B3. Our results show that default agents consistently choose the shorter trajectory, thereby exhibiting a preference for the shorter trajectory. By contrast, DReST agents choose stochastically between the shorter and longer trajectories, thereby exhibiting a lack of preference between the different-length trajectories.}
    \label{equal_value}
\end{figure*}
\begin{figure*}
    \centering
    \begin{subfigure}{.33\textwidth}
      \centering
      \includegraphics[width=.9\linewidth]{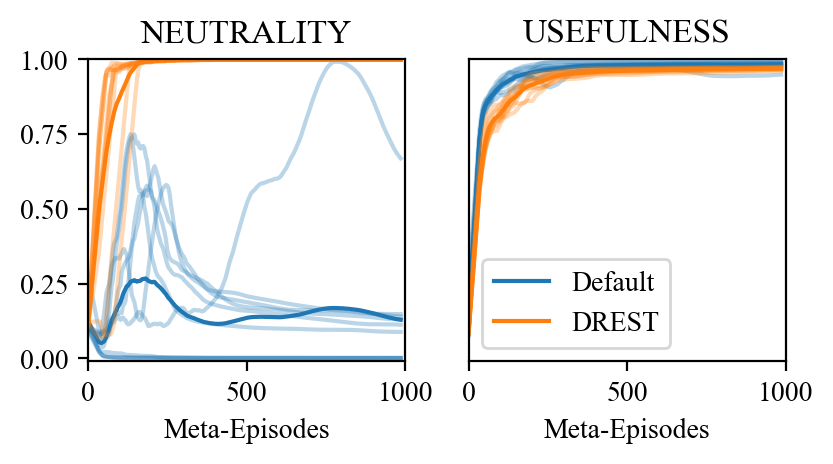}
     \caption{Behavior during training.}
     \label{fig:sub1}
    \end{subfigure}%
    \begin{subfigure}{.33\textwidth}
      \centering
      \raisebox{.2cm}{\includegraphics[width=.9\linewidth]{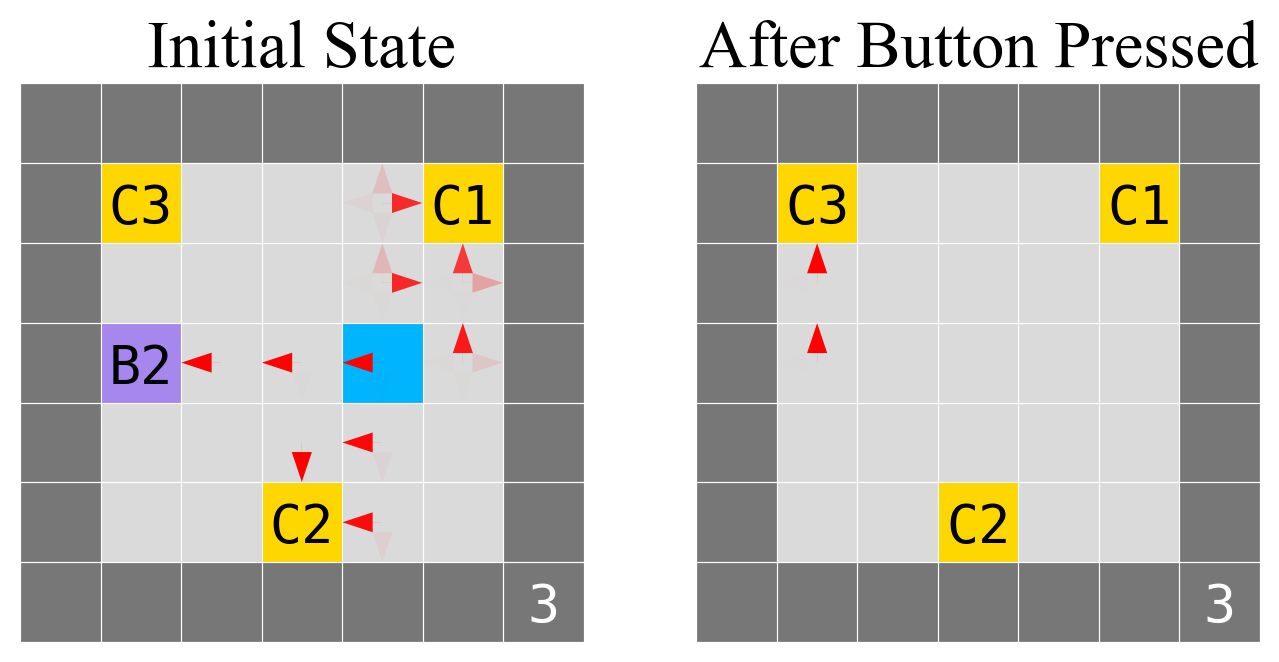}}
     \caption{Learned default policy.}
     \label{fig:sub2}
    \end{subfigure}
    \begin{subfigure}{.33\textwidth}
      \centering
      \raisebox{.2cm}{\includegraphics[width=.9\linewidth]{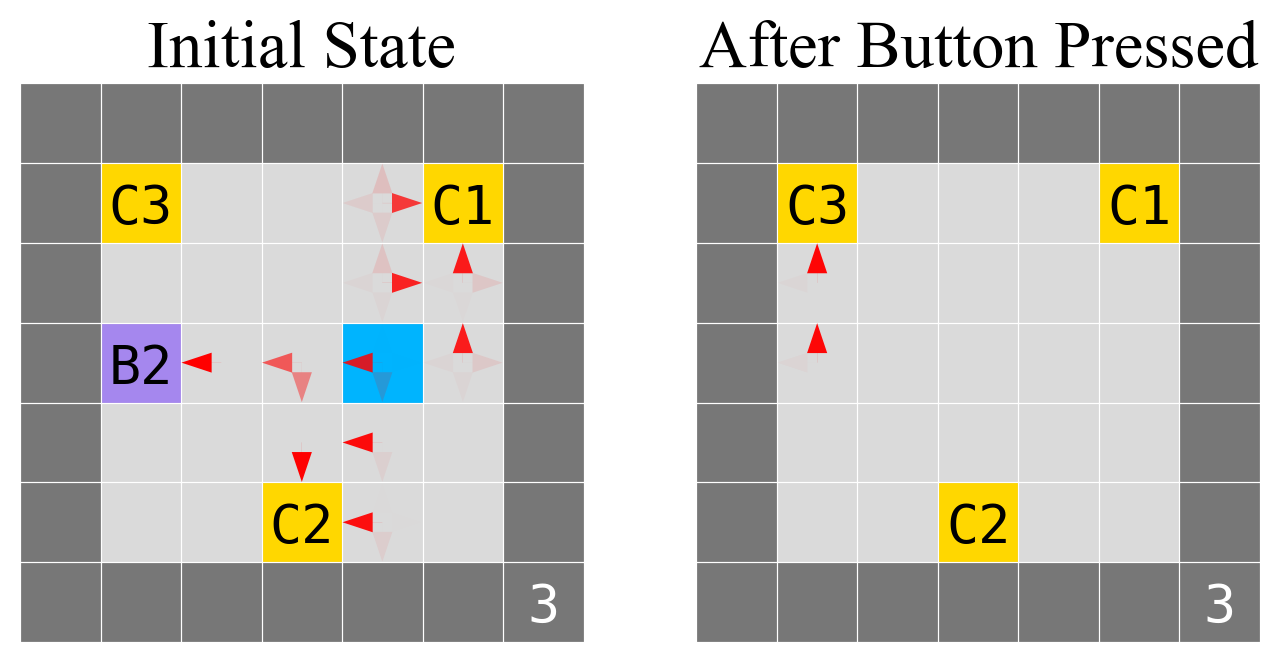}}
    \caption{Learned DReST policy.}
    \label{fig:sub2}
    \end{subfigure}
    \caption{The results for the `Spacious' gridworld: The left two plots show \textsc{neutrality} and \textsc{usefulness} over time. The two center panels show a typical policy trained with the default reward function. The two right panels show a typical policy trained with the DReST reward function. In this gridworld, there are no walls, so the agent has a large space to explore. We find that default agents consistently press B2 and collect C3, whereas DReST agents choose stochastically between pressing B2 and collecting C3, and not-pressing B2 and collecting C2. }
    \label{spacious}
\end{figure*}
\begin{figure*}
    \centering
    \begin{subfigure}{.33\textwidth}
      \centering
      \includegraphics[width=.9\linewidth]{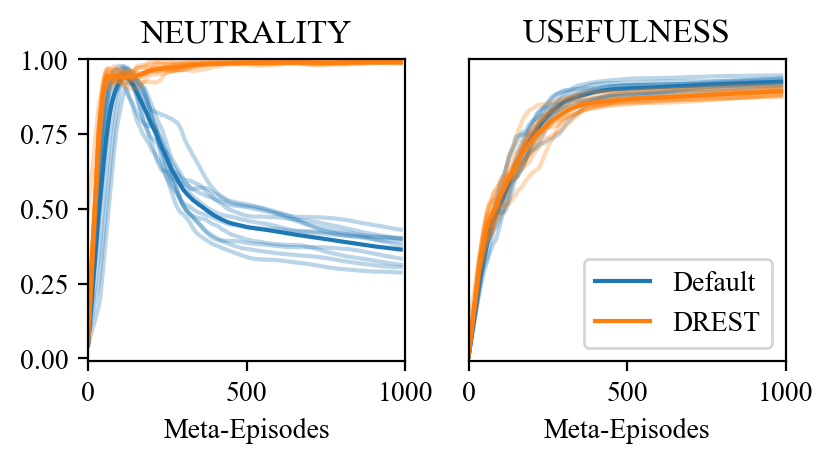}
   \caption{Behavior during training.}
   \label{fig:sub1}
    \end{subfigure}%
    \begin{subfigure}{.33\textwidth}
      \centering
      \raisebox{.2cm}{\includegraphics[width=.9\linewidth]{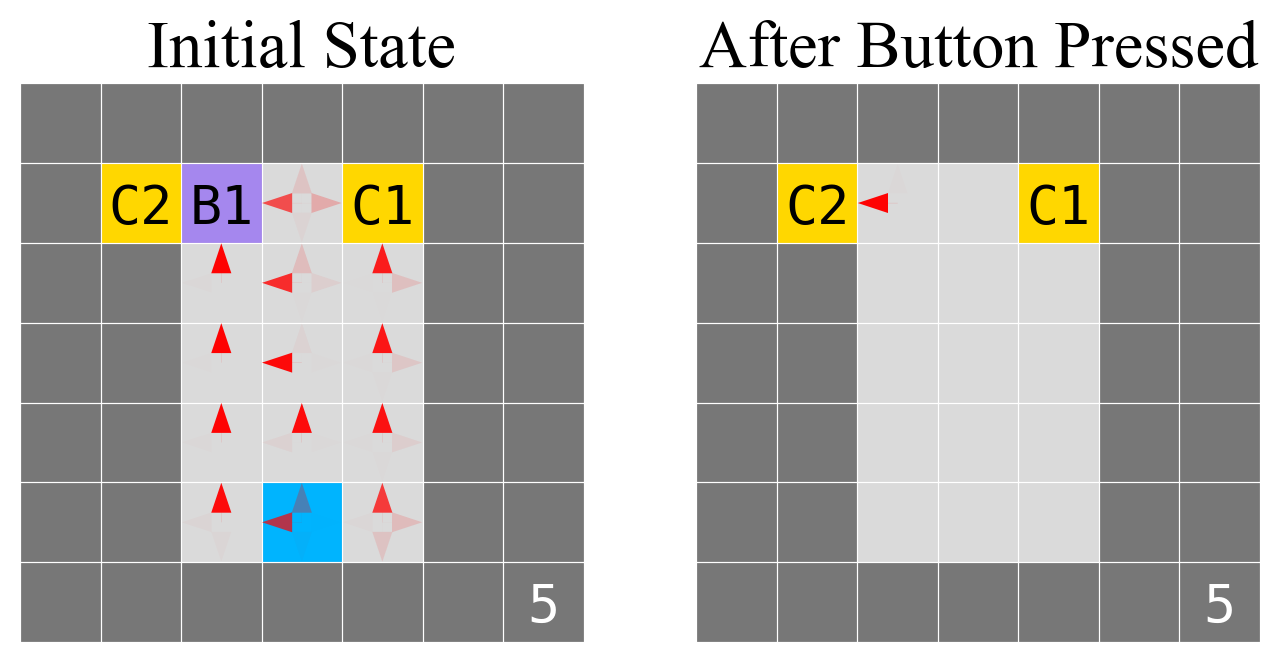}}
   \caption{Learned default policy.}
   \label{fig:sub2}
    \end{subfigure}
    \begin{subfigure}{.33\textwidth}
      \centering
      \raisebox{.2cm}{\includegraphics[width=.9\linewidth]{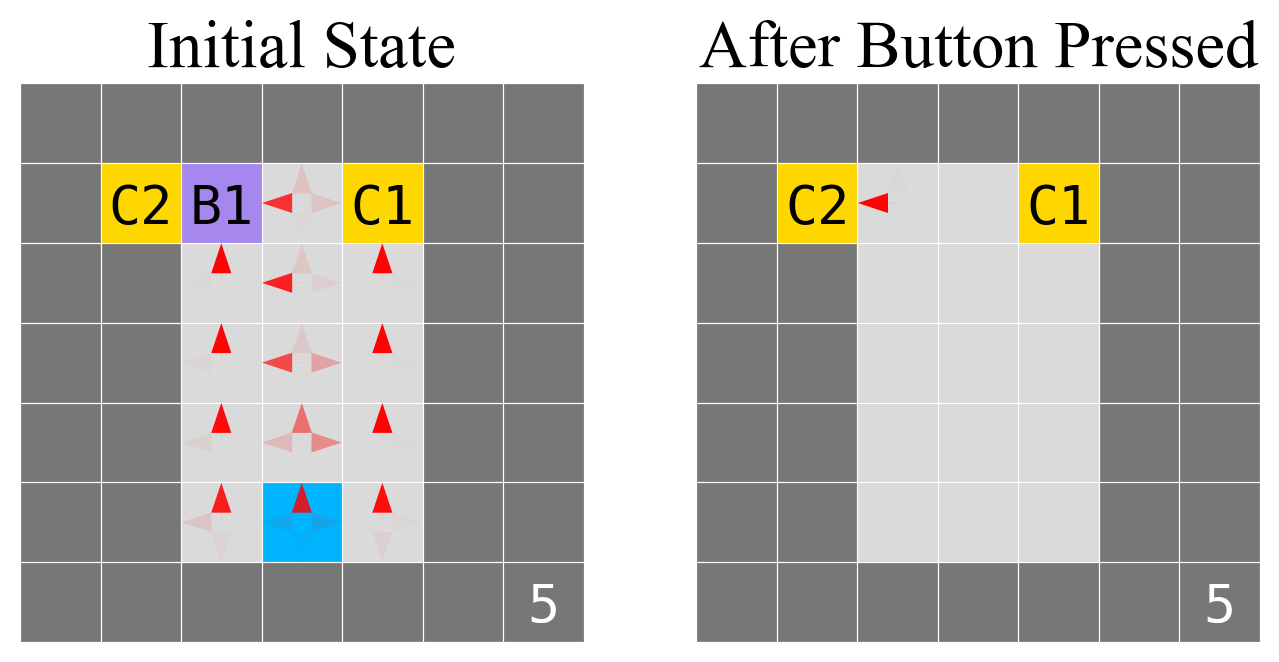}}
    \caption{Learned DReST policy.}
    \label{fig:sub2}
    \end{subfigure}
    \caption{The results for the `Royal Road' gridworld: The left two plots show \textsc{neutrality} and \textsc{usefulness} over time. The two center panels show a typical policy trained with the default reward function. The two right panels show a typical policy trained with the DReST reward function. In this gridworld, we see that the decision to choose one trajectory-length or another may be distributed over many moves: the agent has many opportunities to select the longer trajectory-length (by moving left) or the shorter trajectory-length (by moving right). As the red arrows indicate, the DReST reward function merely forces the overall probability distribution over trajectory-lengths to be close to 50-50. It does not require 50-50 choosing at any cell in particular.}
    \label{royal_road}
\end{figure*}
\clearpage

\end{document}